\documentclass{article}
\usepackage{microtype}
\usepackage{graphicx}
\usepackage{subcaption}
\usepackage{booktabs} % for professional tables
\usepackage{multirow}
\usepackage{pifont}
\usepackage{bbding}
\usepackage{tabularx} % 在导言区添加
\usepackage{hyperref}
\usepackage{tabularx}
\usepackage{booktabs}
\usepackage{multirow}
\usepackage{array}

\usepackage[preprint]{icml2026}

\usepackage{amsmath}
\usepackage{amssymb}
\usepackage{mathtools}
\usepackage{amsthm}
\usepackage{etoolbox}
\usepackage{enumitem}

\usepackage{booktabs}
\usepackage{multirow}
\usepackage{graphicx}   % for \rotatebox
\usepackage{makecell}   % for \makecell
\usepackage{siunitx}    % for \num
\usepackage{graphicx}

\usepackage[capitalize,noabbrev]{cleveref}

\theoremstyle{plain}
\newtheorem{theorem}{Theorem}[section]

\theoremstyle{definition}
\newtheorem{definition}[theorem]{Definition}

\theoremstyle{remark}

\usepackage[textsize=tiny]{todonotes}

\icmltitlerunning{Neuro-Symbolic Experience Replay}

\begin{document}

\twocolumn[
  % \icmltitle{Neuro-Symbolic Experience Replay}
  \icmltitle{From Passive Reuse to Active Reasoning: Grounding Large Language Models for Neuro-Symbolic Experience Replay}



  \icmlsetsymbol{equal}{*}

  \begin{icmlauthorlist}
    \icmlauthor{Yanan Xiao}{equal,yyy}
    \icmlauthor{Yixiang Tang}{equal,comp}
    \icmlauthor{Zechen Feng}{yyy}
    \icmlauthor{Lu Jiang}{sch}
    \icmlauthor{Minghao Yin}{yyy}
    \icmlauthor{Pengyang Wang}{comp}
  \end{icmlauthorlist}

  \icmlaffiliation{yyy}{Affiliation 1}
  \icmlaffiliation{comp}{Affiliation 2}
  \icmlaffiliation{sch}{Affiliation 3}

  \icmlcorrespondingauthor{Pengyang Wang}{pywang@um.edu.mo}

  \icmlkeywords{Machine Learning, ICML}

  \vskip 0.3in
]



\printAffiliationsAndNotice{\icmlEqualContribution}

% re yanan
\begin{abstract}

While experience replay is essential for data efficiency in reinforcement learning (RL), standard methods treat the replay buffer as a passive memory system, prioritizing samples based on numerical prediction errors rather than their semantic significance. 
This approach stands in contrast to human learning, which accelerates mastery by actively abstracting fragmented experiences into behavioral rules. 
To bridge this gap, we propose \textbf{Neuro-Symbolic Experience Replay (NSER)}, a framework that transforms experience replay from a passive sample reuse mechanism into an active engine for knowledge construction. 
Specifically, NSER addresses the incompatibility between linguistic reasoning and numerical optimization through a novel neuro-symbolic grounding pipeline. 
It leverages Large Language Models (LLMs) in a zero-shot manner to induce candidate behavioral rules from accumulated trajectories, grounds these insights into differentiable first-order logic representations, and utilizes the resulting symbolic structures to dynamically reweight the replay distribution. 
By allowing abstract knowledge to directly shape policy optimization, NSER achieves consistent superior sample efficiency and convergence speed across reactive, rule-based, and procedural benchmarks.
Code will be made publicly available upon publication.

% Yanan
% Humans learn efficiently by accumulating experience and abstracting behavioral regularities.
% In contrast, experience replay in reinforcement learning (RL) passively reuses self-generated samples stored in the replay buffer, without the ability to recognize latent behavioral patterns within them.
% As a result, when the environment changes, the agent often exhibits slow adaptation, low sample efficiency, and limited generalization.
% To address this limitation, we propose a \textbf{\underline{N}}euro-\textbf{\underline{S}}ymbolic \textbf{\underline{E}}xperience \textbf{\underline{R}}eplay (NSER) framework.
% It transforms experience replay from passive sample reuse into an active process of symbolic knowledge construction.
% NSER leverages large language models (LLMs) with zero-shot prompting to induce candidate behavioral rules from accumulated experience.
% Since the induced rules are expressed in natural language and are not directly executable, we ground them into differentiable first-order logic representations. 
% These representations produce behavior-guided importance signals that reweight replay trajectories and influencing policy learning. 
% Extensive experiments across multiple benchmark tasks demonstrate that NSER achieves superior overall performance, with consistent improvements in convergence behavior and learning efficiency. 
% Code will be made publicly available upon publication.

\end{abstract}

\section{Introduction}
Experience replay is a cornerstone of modern off-policy reinforcement learning (RL)~\cite{neves2024advances}, enabling agents to reuse collected data for stable and efficient training. 
To maximize the utility of stored data, existing methods often prioritize transitions based on numerical learning signals, such as temporal-difference errors or return-based weighting~\cite{hayes2021replay}. Despite their empirical success, these approaches effectively treat the replay buffer as a \textbf{passive memory mechanism}~\cite{fedus2020revisiting}. 
By viewing stored transitions as isolated and homogeneous samples~\cite{chaudhari2025rlhf}, standard replay strategies focus on short-term error reduction but lack the ability to recognize or exploit the latent behavioral regularities that naturally emerge from accumulated interaction.

\begin{figure}[!t]
\centering
\begin{subfigure}{.48\textwidth}
  \centering
  \includegraphics[width=\linewidth]{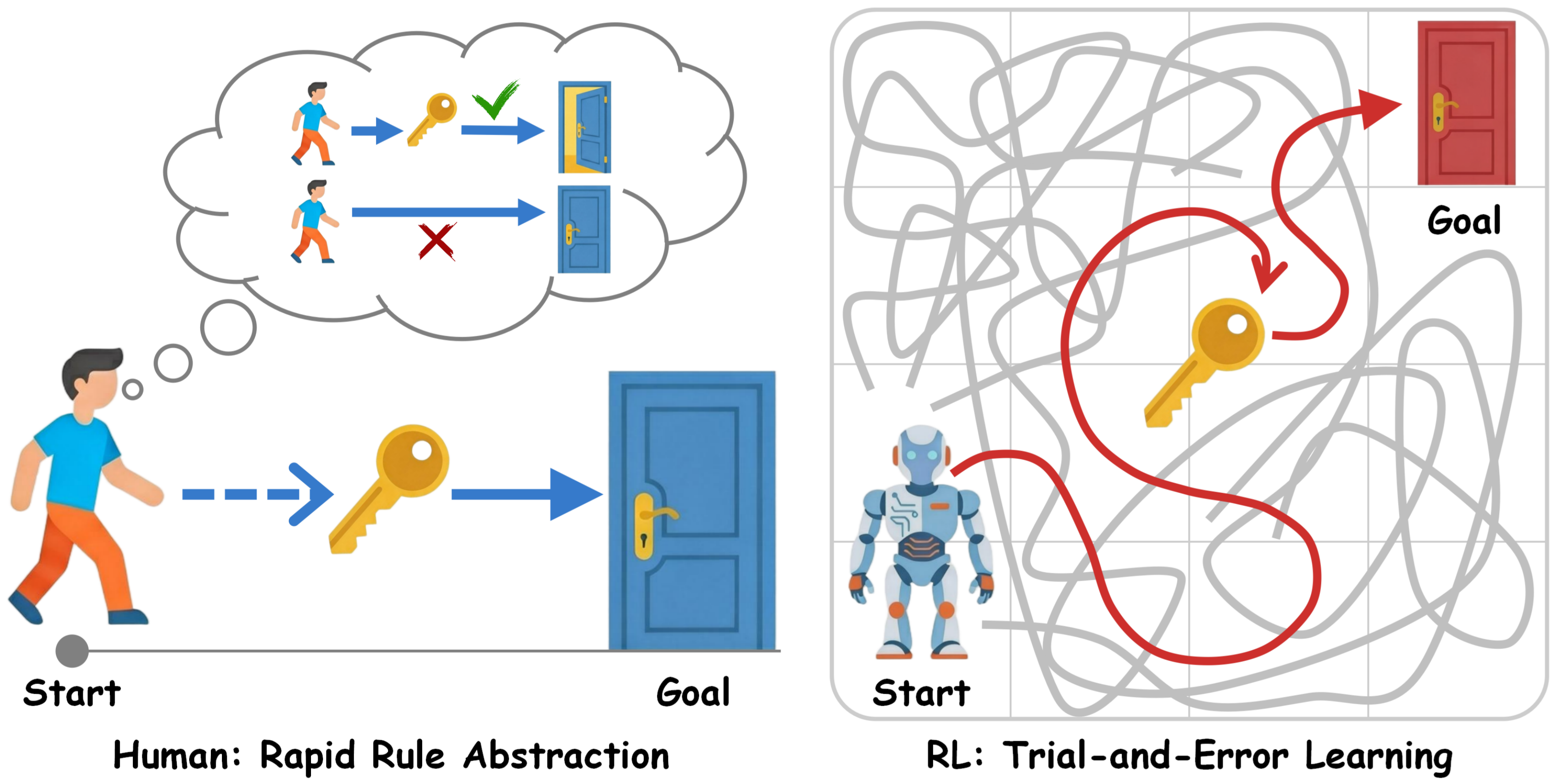}
\end{subfigure}%
\caption{
Illustration of the difference between human learning and reinforcement learning. 
Humans rapidly abstract behavioral effective patterns from limited experience, while reinforcement learning depends on extensive trial-and-error process.
}
\label{fig:1}
\end{figure}

This limitation stands in sharp contrast to human learning~\cite{Volodymyr2015Human}. 
As illustrated in Figure~\ref{fig:1}, humans do not merely store raw experiences~\cite{poddar2024personalizing}; they actively abstract fragmented interactions into behavioral rules: such as recognizing that ``a key is needed to open the door'' to guide future decisions. 
While neuro-symbolic learning offers a pathway to encode such abstract knowledge~\cite{mezzi2025neurosymbolic}, traditional approaches have been constrained by a reliance on predefined predicate spaces or handcrafted rule templates, restricting their flexibility in complex, unknown environments.

Recent advances in Large Language Models (LLMs) provide a powerful alternative for extracting semantic patterns from raw experience~\cite{zhou2025llm}. 
LLMs can process context-rich trajectories to identify high-level strategies. 
However, a fundamental \textbf{grounding gap} remains: the behavioral insights derived by LLMs are expressed in natural language and are not directly executable within the mathematical framework of RL optimization. 
Consequently, this rich semantic knowledge remains inaccessible to the experience replay process~\cite{malviya2022experience}, leaving the agent to rely solely on trial-and-error learning.

To bridge this gap, we propose \textbf{Neuro-Symbolic Experience Replay (NSER)}, a framework that transforms the replay buffer from a passive storage unit into an active engine for knowledge construction. 
Specifically, NSER addresses the disconnect between linguistic abstraction and policy optimization through a novel three-stage pipeline. 
First, it utilizes LLMs in a zero-shot manner to actively induce candidate behavioral rules from accumulated trajectories. 
Second, to make these rules compatible with gradient-based learning, NSER grounds them into executable, differentiable first-order logic (FOL) representations. 
Finally, these grounded rules generate structure-aware importance signals that reweight the replay distribution, ensuring that the policy is shaped by high-quality behavioral patterns.

Our contributions are summarized as follows:
\begin{itemize}
\item We propose a novel replay mechanism that fundamentally redefines experience replay, shifting it from simple sample reuse to an active reasoning process that distills abstract knowledge from past interactions.
\item We introduce a neuro-symbolic grounding mechanism that converts natural language behavioral descriptions into executable, differentiable FOL rules, enabling semantic insights derived from reasoning processes to directly guide numerical policy optimization.
\item Extensive experiments across reactive, rule-based, and procedural benchmarks demonstrate that NSER achieves superior convergence speed and sample efficiency compared to state-of-the-art replay strategies.
\end{itemize}

\section{Preliminaries}
\begin{definition}[Reinforcement Learning Setting]
A reinforcement learning problem is formulated as a Markov Decision Process (MDP)~\cite{lauri2022partially}
$\mathcal{M} = (\mathcal{S}, \mathcal{A}, \mathcal{P}, r, \gamma)$,
where $s \in \mathcal{S}$ and $a \in \mathcal{A}$ denote the state and action spaces,
$\mathcal{P}(s' \mid s,a)$ is the transition probability,
$r(s,a)$ is the reward function, and $\gamma \in [0, 1)$ is the discount factor.
At time step $t$, the environment evolves according to $s_{t+1} \sim \mathcal{P}(\cdot \mid s_t, a_t)$, and the agent receives reward $r_t = r(s_t, a_t)$.
The objective is to learn an optimal policy $\pi(a \mid s)$ that maximizes the expected cumulative discounted return $J(\pi)$, defined as:
\begin{equation}
J(\pi) = \mathbb{E}_{\tau \sim p(\tau \mid \pi)} \left[ \sum_{t=0}^{T} \gamma^t r_t \right].
\end{equation}
\end{definition}

\begin{definition}[Passive Experience Replay]
An interaction trajectory of length $T$ is defined as $\tau = (s_0, a_0, r_0, \dots, s_T)$, representing a complete rollout generated during agent--environment interaction. Given a policy $\pi$, the resulting trajectory distribution is given by:
\begin{equation}
p(\tau \mid \pi) = \rho(s_0) \prod_{t=0}^{T-1} \pi(a_t \mid s_t) \mathcal{P}(s_{t+1} \mid s_t, a_t),
\end{equation}
where $\rho(s_0)$ denotes the initial-state distribution. An experience replay buffer $\mathcal{D}=\{\tau_i\}_{i=1}^N$ stores trajectories collected during interaction. Standard methods treat $\mathcal{D}$ as a \textbf{passive memory system}, typically reweighting the sampling distribution $q(\tau)$ based on local numerical signals like TD-error rather than semantic significance.
\end{definition}

\begin{definition}[Grounded Behavioral Rule]
A symbolic behavioral rule $\psi$ is defined as a first-order logic (FOL) expression capturing prerequisite or constraint relations between states and actions~\cite{lalwani2025autoformalizing}. For a given decision context $x$, a rule follows the structure:
\begin{equation}
\psi: \forall x, \mathcal{C}(x) \Rightarrow \mathcal{E}(x),
\end{equation}
where $\mathcal{C}(\cdot)$ and $\mathcal{E}(\cdot)$ denote prerequisite conditions and induced behavioral outcomes, respectively. To enable numerical optimization, we define a \textbf{grounding function} $f_\psi(\tau) \in [0, 1]$ that maps a trajectory $\tau$ to a differentiable satisfaction degree. This allows abstract logic to serve as structured signals for reweighting the replay distribution to improve sample efficiency during policy training.
\end{definition}

\subsection{Problem Formulation}
We reformulate the replay buffer as an active knowledge engine by inducing symbolic rules $\mathcal{Z} = \{\psi_k\}_{k=1}^K$ to characterize latent regularities in the trajectory distribution $p(\tau)$. We define an \textbf{active induction mechanism} $\mathcal{I}$ that leverages LLMs to distill these relations:
\begin{equation}
\mathcal{Z} = \mathcal{I}(\mathcal{D}), \quad \text{where } \psi_k \sim \text{LLM}(\tau) \text{ for } \tau \in \mathcal{D}.
\end{equation}
To bridge the grounding gap, these linguistic insights are mapped to a structure-aware distribution $\tilde{p}(\tau)$. The relationship between the original replay distribution and the NSER-guided distribution is defined as:
\begin{equation}
\tilde{p}(\tau) = \frac{p(\tau) \cdot \exp(\mathcal{G}(\tau, \mathcal{Z}))}{\sum_{\tau' \in \mathcal{D}} p(\tau') \cdot \exp(\mathcal{G}(\tau', \mathcal{Z}))},
\end{equation}
where $\mathcal{G}(\tau, \mathcal{Z})$ is the grounding function measuring the satisfaction of induced rules. By prioritizing trajectories with high semantic significance, NSER enables abstract knowledge to directly guide the policy optimization process.

\begin{figure*}[!t]
\centering
\includegraphics[width=0.95\textwidth]{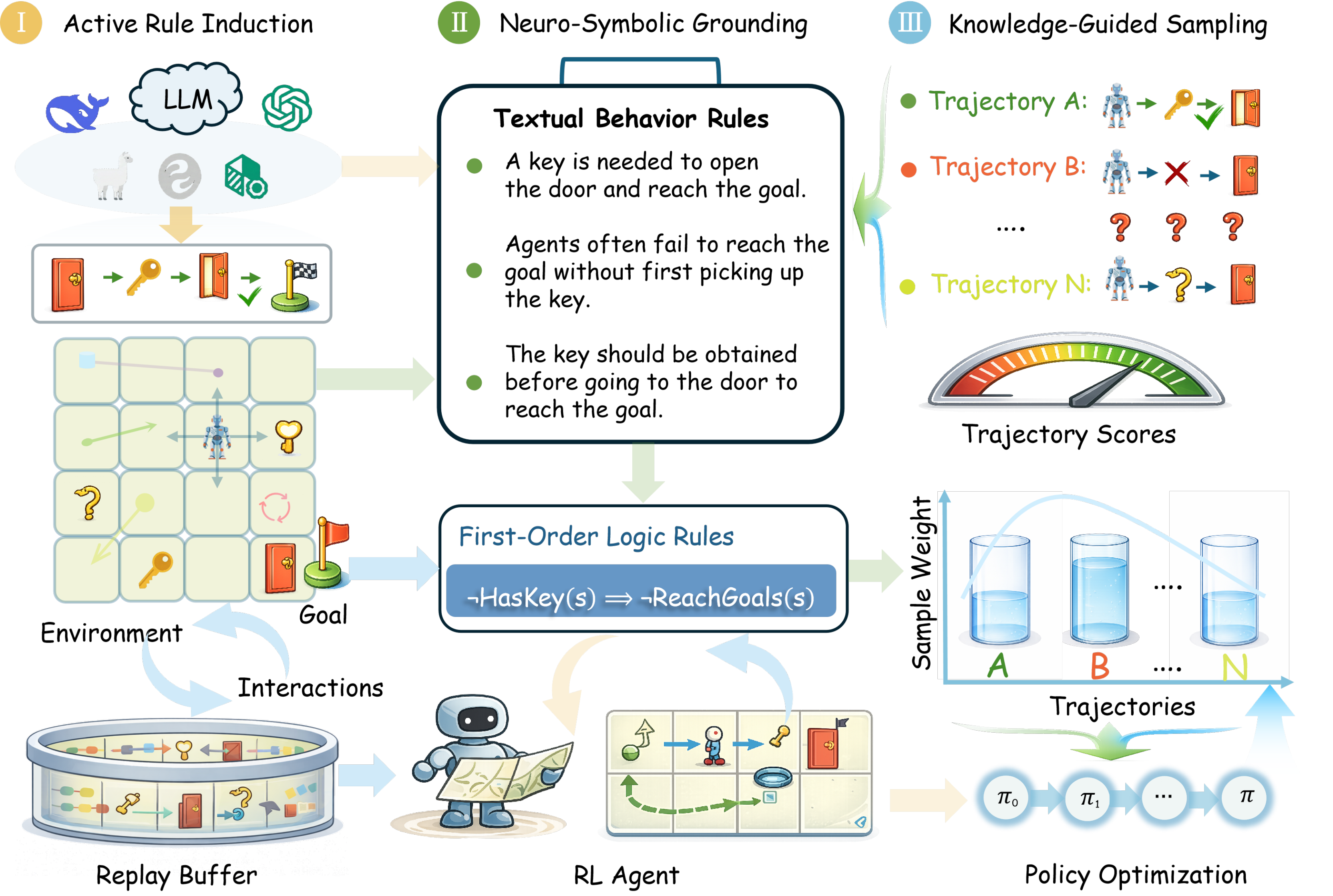}
\caption{
Overview of the NSER framework. 
Starting from the environment interaction, raw trajectories are stored in a replay buffer. 
Stage i involves active rule induction, where an LLM distills behavioral logic from serialized experiences. 
Stage ii represents neuro-symbolic grounding, converting these insights into logical rules and differentiable predicates. Stage iii shows knowledge-guided sampling, where satisfaction scores reshape the replay distribution to prioritize semantically significant data for policy optimization.
}
\label{fig:2}
\end{figure*}

\section{Neuro-Symbolic Experience Replay}
As illustrated in Figure~\ref{fig:2}, NSER transforms passive sample reuse into an active knowledge construction process through three synergistic stages: 
(i) \textbf{Active Rule Induction}, utilizing zero-shot LLMs to distill abstract behavioral regularities from trajectories; 
(ii) \textbf{Neuro-Symbolic Grounding}, converting linguistic insights into differentiable FOL representations to bridge the grounding gap; 
and (iii) \textbf{Knowledge-Guided Sampling}, leveraging grounded rules to reweight the replay distribution for policy optimization.

\subsection{Active Rule Induction}

In this section, we detail the process of distilling latent behavioral regularities from raw trajectories. To ensure the entire induction is \textbf{active} and \textbf{autonomous}, we leverage zero-shot LLMs to analyze experiences without predefined behavioral templates, followed by a representation alignment stage to ensure semantic stability.

\noindent \textbf{Fixed-Template Serialization.}
To enable LLMs to reason over numerical interaction data, we first convert each trajectory $\tau=(s_0,a_0,r_0,\ldots,s_T)$ into a serialized narrative $x(\tau)$. 
We employ a fixed descriptive schema $\psi(\cdot)$ that maps raw state-action vectors to semantic labels, preserving their temporal dependency, defined as
\begin{equation}
x(\tau) \triangleq \big[\psi_s(s_0), \psi_a(a_0), \psi_r(r_0), \dots, \psi_s(s_T)\big].
\end{equation}
By verbalizing numerical values into a structured format, we provide a readable context that allows the LLM to identify logical transitions within the episode.

\noindent \textbf{Zero-Shot Pattern Proposal.}
Given $x(\tau)$, we leverage the LLM as a zero-shot engine to induce candidate rules by identifying latent regularities implied by the trajectory outcome. Instead of prescribing behaviors, the LLM autonomously discovers critical bottlenecks, such as missing prerequisite items, without task-specific supervision. To capture these insights, we decode $M$ diverse linguistic proposals:
\begin{equation}
\mathcal{U}(\tau) = \{u^{(m)}\}_{m=1}^{M}, \quad u^{(m)} \sim p_{\mathrm{LM}}(\cdot \mid x(\tau)),
\end{equation}
where $\mathcal{U}(\tau)$ represents the set of diverse linguistic proposals that capture the underlying behavioral logic of the trajectory $\tau$ from multiple semantic perspectives.

\noindent \textbf{Semantic Consensus Alignment.}
To achieve a stable semantic consensus, we align the proposals $\mathcal{U}(\tau)$ with a set of learnable prototypes $\{c_k\}_{k=1}^K$ in a shared embedding space. Each prototype $c_k \in \mathbb{R}^d$ represents a distinct logic $z_k$. We map diverse proposals to these prototypes via a text encoder $\varphi(\cdot)$ and a max-pooling alignment mechanism:
\begin{equation}
q(z_k \mid \tau) = \frac{\exp\left( \beta \max_{u \in \mathcal{U}(\tau)} \langle \varphi(u), c_k \rangle \right)}{\sum_{j=1}^K \exp\left( \beta \max_{u \in \mathcal{U}(\tau)} \langle \varphi(u), c_j \rangle \right)},
\end{equation}
where $\beta$ is a temperature parameter. To ensure semantic consistency, prototypes are optimized by maximizing agreement over the buffer $\mathcal{D}$, defined as
\begin{equation}
\max_{\{c_k\}} \sum_{\tau \in \mathcal{D}} \log \sum_{k=1}^K \exp\left( \beta \max_{u \in \mathcal{U}(\tau)} \langle \varphi(u), c_k \rangle \right).
\end{equation}
The resulting assignment $\Phi(\tau) = \arg\max_{z} q(z \mid \tau)$ effectively transforms noisy linguistic outputs into stable and interpretable symbolic relations $\mathcal{Z}$.

\subsection{Neuro-Symbolic Grounding}
NSER transforms induced behavioral relations into executable symbolic structures, enabling linguistic insights to be evaluated on continuous trajectories. This is achieved via a differentiable reasoning pipeline that maps discrete logical rules into a continuous optimization landscape.

\noindent \textbf{Symbolic Rule Construction.}
Each behavioral relation $z_k \in \mathcal{Z}$ is modeled as a logical implication that characterizes its prerequisite structure. Specifically, we represent the relation as a mapping from a set of observed state-action conditions to a specific behavioral outcome:
\begin{equation}
z_k :\quad \bigwedge_{\ell=1}^{L_k} \mathcal{P}_k^{(\ell)} \;\Rightarrow\; \mathcal{Q}_k,
\end{equation}
where $\bigwedge$ denotes the logical conjunction of $L_k$ symbolic predicates $\mathcal{P}_k^{(\ell)}$ describing behavioral conditions, and $\mathcal{Q}_k$ denotes the induced outcome.
This structure transforms interactions into structured, behavioral knowledge.

\noindent \textbf{Differentiable Predicate Evaluation.}
To ensure compatibility with gradient-based optimization, we ground each predicate $\mathcal{P}$ into a learnable scoring function $E_{\mathcal{P}}(\cdot)$ over the trajectory space. The predicate is computed as:
\begin{equation}
\mu_{\mathcal{P}}(\tau) = \sigma\!\left( E_{\mathcal{P}}(\tau) \right) \in [0, 1],
\end{equation}
where $\sigma(\cdot)$ is the sigmoid function. We employ a differentiable product $t$-norm to represent the conjunction $\bigwedge$ as a soft satisfaction degree, defined as
\begin{equation}
\mu_{z_k}(\tau) = \prod_{\ell=1}^{L_k} \mu_{\mathcal{P}_k^{(\ell)}} (\tau).
\end{equation}
where the product $\prod$ aggregates activations into a joint satisfaction degree, enabling discrete logic to interface directly with the policy gradient during replay.

\subsection{Knowledge-Guided Sampling}
NSER leverages grounded knowledge to transform passive replay into an active engine by reweighting the sampling distribution based on semantic significance.

\noindent \textbf{Relation-Based Trajectory Scoring.}
For each trajectory $\tau \in \mathcal{D}$, we compute a structure-aware importance score $w(\tau)$ to emphasize alignment with distilled knowledge:
\begin{equation}
w(\tau) = \sum_{k=1}^{K} \mu_{z_k}(\tau)^p, \quad p \ge 1,
\end{equation}
where $p$ is a sensitivity parameter. This scoring mechanism ensures that transitions consistent with symbolic rules are prioritized during the training cycle.

\noindent \textbf{Structure-Aware Replay Distribution.}
The knowledge-guided distribution $\tilde{p}(\tau)$ is formulated via a Softmax transformation to balance exploration with the exploitation of symbolic insights.
The formulation is defined as
\begin{equation}
\tilde{p}(\tau) = \frac{\exp\left( \eta \, w(\tau) \right)}{\sum_{\tau' \in \mathcal{D}} \exp\left( \eta \, w(\tau') \right)},
\end{equation}
where $\eta$ controls prioritization intensity. This strategically focuses agent attention on high-quality interactions that reinforce induced behavioral regularities.

\subsection{Knowledge-Infused Learning Objective}
The optimization objective couples symbolic reasoning with gradient-based learning. Let $\mathcal{L}_{\mathrm{RL}}(\tau;\theta)$ denote the standard reinforcement learning loss.
The training objectives of NSER can be formally defined as:
\begin{equation}
\mathcal{L}(\theta) = \mathbb{E}_{\tau \sim \tilde{p}(\tau)} \left[ \mathcal{L}_{\mathrm{RL}}(\tau;\theta) \right].
\end{equation}
By prioritizing semantically significant data, the resulting gradient update is reformulated as:
\begin{equation}
\nabla_{\theta}\mathcal{L}(\theta) = \mathbb{E}_{\tau \sim \tilde{p}(\tau)} \left[ \nabla_{\theta} \mathcal{L}_{\mathrm{RL}}(\tau;\theta) \right].
\end{equation}
In this manner, grounded knowledge shapes gradient estimation, steering the RL agent toward logical regularities distilled from interaction experience.

\begin{table*}[t]
\centering
\small
\setlength{\tabcolsep}{3pt}
\renewcommand{\arraystretch}{1.2}
\caption{
Performance comparison between uniform experience replay (UER) and the proposed NSER
across three off-policy algorithms (DQN, C51, and QR-DQN) evaluated on reactive, rule-based,
and procedural environments.
Results are reported as \textbf{UER/NSER}, where performance is measured by episodic return
and area under the learning curve (AUC), efficiency by Steps-to-$\tau$ and $N_{\mathrm{conv}}$,
and computational cost by the relative speedup ratio $\eta = T_{\mathrm{UER}} / T_{\mathrm{NSER}}$.
Bold values indicate the better result.
}
\begin{tabular}{c c c c c c c c}
\toprule
\multirow{2}{*}{} & \multirow{2}{*}{\textbf{Env Type}} & \multirow{2}{*}{\textbf{Environment}} &
\multicolumn{2}{c}{\textbf{Performance}} &
\multicolumn{2}{c}{\textbf{Efficiency}} &
\multicolumn{1}{c}{\textbf{Cost}} \\
\cmidrule(lr){4-5}\cmidrule(lr){6-7}\cmidrule(lr){8-8}
& & & Return $\uparrow$  & AUC $\uparrow$  &
Steps-to-$\tau$ $\downarrow$  & $N_{\mathrm{conv}}$ $\downarrow$  &
$\eta$ $\uparrow$  \\
\midrule
\multirow{6}{*}{\rotatebox[origin=c]{90}{DQN}}
& \multirow{2}{*}{Reactive}
& Acrobot-v1 
& $-123 \!\pm\! 88$/$\mathbf{-112\!\pm\!77}$ & $-2.71\times 10^{5}$/$\mathbf{-2.82\times 10^{5}}$ & 154k/\textbf{99k} & 271k/\textbf{14k} & \textbf{1.06} \\
& 
& CartPole-v1 
& $470\!\pm\!66$/$\mathbf{500\!\pm\!14}$ & $3.30\times 10^{5}$/$\mathbf{3.43\times 10^{5}}$ & 19k/\textbf{14k} & 174k/\textbf{129k} & \textbf{2.63} \\
\cmidrule(lr){2-8}
& \multirow{2}{*}{Rule}
& FrozenLake-v1 
& $0.04\!\pm\!0.35$/$\mathbf{0.11\!\pm\!0.31}$ & $56$/$\mathbf{131}$ & 1.7k/\textbf{0.03k} & 19k/\textbf{1.5k} & \textbf{1.55} \\
& 
& Taxi-v3 
& $-452\!\pm\!222$/$\mathbf{-428\!\pm\!263}$ & $-7.91\times 10^{5}$/$\mathbf{-6.32\times 10^{5}}$ & 200k/\textbf{199k} & 264k/\textbf{179k} & \textbf{1.04} \\
\cmidrule(lr){2-8}
& \multirow{2}{*}{Procedural}
& Procgen-CoinRun 
& $1.2\!\pm\!3.2$/$\mathbf{5.65\!\pm\!3.71}$ & $2.83\times 10^{3}$/$\mathbf{3.0\times 10^{5}}$ & 36k/\textbf{21k} & 532k/\textbf{50k} & \textbf{2.64} \\
& 
& Procgen-Maze 
& $1.1\!\pm\!3.1$/$\mathbf{5.0\!\pm\!5.0}$ & $2.26\times 10^{3}$/$\mathbf{4.62\times 10^{3}}$ & 586/\textbf{19} & 416k/\textbf{6k} & \textbf{1.43} \\
\midrule
\multirow{6}{*}{\rotatebox[origin=c]{90}{C51}}
& \multirow{2}{*}{Reactive}
& Acrobot-v1 
& $-196\!\pm\!27$/$\mathbf{-142\!\pm\!105}$ & $\-1.86\times 10^{5}$/$\mathbf{-1.73\times 10^{5}}$ & 87k/\textbf{11k} & 188k/\textbf{1k} & \textbf{1.02} \\
& 
& CartPole-v1 
& $11\!\pm\!8$/$\mathbf{163\!\pm\!100}$ & $1.52\times 10^{5}$/$\mathbf{1.91\times 10^{5}}$ & 3k/\textbf{1.4k} & 151k/\textbf{101k} & \textbf{1.68} \\
\cmidrule(lr){2-8}
& \multirow{2}{*}{Rule}
& FrozenLake-v1 
& $0.04\!\pm\!0.02$/$\mathbf{0.37\!\pm\!0.48}$ & $8.0$/$\mathbf{203}$ & 2k/\textbf{1k} & 3k/\textbf{1k} & \textbf{1.37} \\
& 
& Taxi-v3 
& $-218\!\pm\!14$/$\mathbf{-187\!\pm\!15}$ & $-9.76\times 10^{5}$/$\mathbf{-8.81\times 10^{5}}$ & 400k/\textbf{398k} & 400k/\textbf{398k} & \textbf{1.13} \\
\cmidrule(lr){2-8}
& \multirow{2}{*}{Procedural}
& Procgen-CoinRun 
& $2.2\!\pm\!4.7$/$\mathbf{2.5\!\pm\!4.3}$ & $6.13\times 10^{3}$/$\mathbf{6.34\times 10^{3}}$ & 30k/\textbf{5k} & 980k/\textbf{876k} & \textbf{1.08} \\
& 
& Procgen-Maze 
& $1.1\!\pm\!3.1$/$\mathbf{5.1\!\pm\!5.0}$ & $3.05\times 10^{3}$/$\mathbf{4.33\times 10^{3}}$ & 2k/\textbf{1k} & 806k/\textbf{1k} & \textbf{1.35} \\
\midrule
\multirow{6}{*}{\rotatebox[origin=c]{90}{QR-DQN}}
& \multirow{2}{*}{Reactive}
& Acrobot-v1 
& $-215\!\pm\!155$/$\mathbf{-91.70\pm\!27.28}$& $\-2.38\times 10^{5}$/$\mathbf{-1.41\times 10^{5}}$ & 48k/\textbf{46k} & 83k/\textbf{1k} & \textbf{1.31} \\
& 
& CartPole-v1 
& $9.4\!\pm\!0.7$/$\mathbf{12.5\!\pm\!0.8}$ & $9.64\times 10^{3}$/$\mathbf{1.18\times 10^{4}}$ & 16k/\textbf{12k} & 15k/\textbf{10k} & \textbf{1.19} \\
\cmidrule(lr){2-8}
& \multirow{2}{*}{Rule}
& FrozenLake-v1 
& $0.06\!\pm\!0.24$/$\mathbf{0.08\!\pm\!0.27}$ & $86.0$/$\mathbf{98.0}$ & 8k/\textbf{1k} & 4k/\textbf{3.5k} & \textbf{1.03} \\
& 
& Taxi-v3 
& $-297\!\pm\!32$/$\mathbf{-207\!\pm\!14}$ & $-9.11\times 10^{5}$/$\mathbf{-8.78\times 10^{5}}$ & 400k/\textbf{398k} & 400k/\textbf{398k} & \textbf{1.19} \\
\cmidrule(lr){2-8}
& \multirow{2}{*}{Procedural}
& Procgen-CoinRun 
& $3.1\!\pm\!2.6$/$\mathbf{4.8\!\pm\!4.5}$ & $1.77\times 10^{3}$/$\mathbf{5.67\times 10^{3}}$ & 7k/\textbf{5k} & 282k/\textbf{205k} & \textbf{1.27} \\
& 
& Procgen-Maze 
& $0.90\!\pm\!2.86$/$\mathbf{5.0\!\pm\!5.0}$ & $710$/$\mathbf{4.62\times 10^{3}}$ & 10k/\textbf{2k} & 41k/\textbf{6k} & \textbf{1.73} \\
\bottomrule
\end{tabular}

\label{tab:1}
\end{table*}

\section{Experiments}
This study conducts a comprehensive experimental evaluation of NSER to investigate the effectiveness.
Specifically, we seek to answer the following research questions:
\begin{itemize}[leftmargin=*, itemsep=2pt, parsep=0pt]

\item \textbf{Q1 (Generalization):} Does NSER generalize effectively across diverse environments and various RL backbones?

\item \textbf{Q2 (Comparative Advantage):} Does NSER offer superior performance improvements compared to existing representative experience replay methods?

\item \textbf{Q3 (Induction Stability):} How does active rule induction design reduce semantic noise and ensure the stability of induced behavioral relations?

\item \textbf{Q4 (Grounding Effect):} To what extent does differentiable neuro-symbolic grounding improve gradient stability compared to discrete logic?

\item \textbf{Q5 (Sampling Efficiency):} How effectively does knowledge-guided sampling accelerate convergence by prioritizing semantically significant experiences?

\item \textbf{Q6 (Interpretability):} What symbolic behavioral patterns are induced during the learning process, and how do they evolve to guide the agent?

\end{itemize}

\begin{table*}[t]
\centering
\small
\setlength{\tabcolsep}{6pt}
\renewcommand{\arraystretch}{1.1}
\caption{
Comparison of alternative experience replay strategies under DQN across different environment.
We compare uniform experience replay (UER), prioritized experience replay (PER), combined experience replay (CER), n-step replay, and the proposed NSER.
Performance, learning efficiency, and computational overhead are reported.
All speedup ratios are defined as $\eta = T_{\mathrm{UER}} / T_{\mathrm{NSER}}$, where $\eta > 1$ indicates faster training than UER.
Bold values denote the best result for each metric within the same environment.
}
\begin{tabular}{c c c c c c}
\toprule
\multirow{2}{*}{Game/Env} &
\multicolumn{2}{c}{Performance} &
\multicolumn{2}{c}{Efficiency} &
Overhead \\
\cmidrule(lr){2-3}\cmidrule(lr){4-5}\cmidrule(lr){6-6}
& Return $\uparrow$ & AUC $\uparrow$ &
Steps-to-$\tau$ $\downarrow$ & $N_{\mathrm{conv}}$ $\downarrow$ &
$\eta $ \\
% \midrule
% \multicolumn{6}{l}{\textbf{Reactive Environments}} \\
\midrule
CartPole-v1 &
\makecell[l]{UER: $470\pm66$\\PER: $480\pm64$\\CER: $89\pm71$\\n-step: $137\pm139$\\NSER: $\mathbf{500\pm14}$} &
\makecell[l]{UER: $3.30\times 10^{5}$\\PER: $3.41\times 10^{5}$\\CER: $3.10\times 10^{5}$\\n-step: $1.30\times 10^{5}$\\NSER: $\mathbf{3.43\times 10^{5}}$} &
\makecell[l]{UER: 19k\\PER: 34k\\CER: 100k\\n-step: 30k\\NSER: \textbf{14k}} &
\makecell[l]{UER: 174k\\PER: 161k\\CER: 310k\\n-step: 137k\\NSER: \textbf{129k}} &
\makecell[l]{UER: 1.00\\PER: 0.33\\CER: 1.04\\n-step: 1.75\\NSER: \textbf{2.63}} \\
\midrule
Acrobot-v1 &
\makecell[l]{UER: $-123\pm88$\\PER: $-122\pm86$\\CER: $-252\pm160$\\n-step: $-114\pm68$\\NSER: $\mathbf{-112\pm77}$} &
\makecell[l]{UER: $-2.71\times 10^{5}$\\PER: $-2.70\times 10^{5}$\\CER: $-2.96\times 10^{5}$\\n-step: $-2.43\times 10^{5}$\\NSER: $\mathbf{-2.82\times 10^{5}}$} &
\makecell[l]{UER: 154k\\PER: 169k\\CER: 175k\\n-step: 163k\\NSER: \textbf{99k}} &
\makecell[l]{UER: 271k\\PER: 267k\\CER: 296k\\n-step: 205k\\NSER: \textbf{14k}} &
\makecell[l]{UER: 1.00\\PER: 0.42\\CER: 0.94\\n-step: 0.96\\NSER: \textbf{1.06}} \\
\midrule
FrozenLake-v1 &
\makecell[l]{UER: $0.04\pm0.35$\\PER: $0.00\pm0.00$\\CER: $0.03\pm0.17$\\n-step: $0.01\pm0.10$\\NSER: $\mathbf{0.11\pm0.31}$} &
\makecell[l]{UER: $56$\\PER: $15$\\CER: $16$\\n-step: $14$\\NSER: $\mathbf{131}$} &
\makecell[l]{UER: 1709\\PER: 301\\CER: 568\\n-step: 903\\NSER: \textbf{32}} &
\makecell[l]{UER: 19k\\PER: 7.7k\\CER: 7.8k\\n-step: 7.7k\\NSER: \textbf{1.5k}} &
\makecell[l]{UER: 1.00\\PER: 0.53\\CER: 0.50\\n-step: 0.34\\NSER: \textbf{1.55}} \\
\midrule
Taxi-v3 &
\makecell[l]{UER: $-452\pm222$\\PER: $-772\pm78$\\CER: $-759\pm101$\\n-step: $-758\pm112$\\NSER: $\mathbf{-428\pm263}$} &
\makecell[l]{UER: $-7.91\times 10^{5}$\\PER: $-7.75\times 10^{5}$\\CER: $-7.74\times 10^{5}$\\n-step: $-7.73\times 10^{5}$\\NSER: $\mathbf{-6.32\times 10^{5}}$} &
\makecell[l]{UER: 200k\\PER: 197k\\CER: 297k\\n-step: 243k\\NSER: \textbf{199k}} &
\makecell[l]{UER: 264k\\PER: 187k\\CER: 189k\\n-step: 191k\\NSER: \textbf{179k}} &
\makecell[l]{UER: 1.00\\PER: 0.74\\CER: 0.52\\n-step: 0.56\\NSER: \textbf{1.04}} \\
\midrule
Procgen-CoinRun &
\makecell[l]{UER: $1.2\pm3.2$\\PER: $3.4\pm4.7$\\CER: $2.4\pm4.3$\\n-step: $3.1\pm4.6$\\NSER: $\mathbf{5.65\pm3.71}$} &
\makecell[l]{UER: $2.83\times10^{4}$\\PER: $2.68\times10^{4}$\\CER: $2.5\times10^{4}$\\n-step: $2.59\times10^{4}$\\NSER: $\mathbf{3.0\times10^{5}}$} &
\makecell[l]{UER: 36k\\PER: 279k\\CER: 316k\\n-step: 346k\\NSER: \textbf{21k}} &
\makecell[l]{UER: 532k\\PER: 658k\\CER: 663k\\n-step: 652k\\NSER: \textbf{50k}} &
\makecell[l]{UER: 1.00\\PER: 0.43\\CER: 0.43\\n-step: 0.52\\NSER: \textbf{2.64}} \\
\midrule
Procgen-Maze &
\makecell[l]{UER: $1.1\pm3.1$\\PER: $4.0\pm4.9$\\CER: $4.4\pm5.0$\\n-step: $3.7\pm4.8$\\NSER: $\mathbf{5.0\pm5.0}$} &
\makecell[l]{UER: $2.26\times10^{4}$\\PER: $4.27\times10^{4}$\\CER: $4.35\times10^{4}$\\n-step: $4.37\times10^{4}$\\NSER: $\mathbf{4.62\times10^{4}}$} &
\makecell[l]{UER: 586\\PER: 571\\CER: 120\\n-step: 3507\\NSER: \textbf{19}} &
\makecell[l]{UER: 416k\\PER: 348k\\CER: 349k\\n-step: 345k\\NSER: \textbf{6k}} &
\makecell[l]{UER: 1.00\\PER: 0.36\\CER: 0.45\\n-step: 0.33\\NSER: \textbf{1.43}} \\
\bottomrule
\end{tabular}
\label{tab:2}
\end{table*}

\subsection{Benchmarks and Environment Settings}
We evaluate NSER across diverse benchmarks~\cite{chevalier2018minimalistic,krishnan2023archgym} categorized into three distinct decision structures. 
(i) \textbf{Reactive environments} (e.g., CartPole-v1, Acrobot-v1) feature stationary dynamics to assess basic learning speed; 
(ii) \textbf{Rule-based environments} (e.g., FrozenLake-v1, Taxi-v3) involve implicit action constraints where valid decisions depend on satisfying specific state-logical conditions; 
and (iii) \textbf{Procedurally generated environments} (e.g., CoinRun, Maze) introduce distinct instances to assess generalization under unseen but structurally similar configurations. 
By spanning these domains, we evaluate NSER's capability to distill and exploit behavioral regularities across varying levels of temporal dependency and logical complexity. 
Detailed environment configurations are provided in Appendix~\ref{app:env}.

\subsection{Baselines}
\noindent \textbf{Off-policy RL Algorithms.}
To assess the general applicability of NSER across different value-based learning paradigms, we integrate our framework with three representative algorithms: 
(i) Deep Q-Network (DQN)~\cite{Volodymyr2015Human}, 
(ii) Categorical DQN (C51)~\cite{2018A}, and 
(iii) Quantile Regression DQN (QR-DQN)~\cite{10.5555/3504035.3504388}. 
These span a diverse spectrum of value formulations, including expectation-based, categorical distributional, and quantile-based representations.

\noindent \textbf{Experience Replay Strategies.}
To isolate the contribution of knowledge-guided sampling, we compare NSER against several prominent replay mechanisms~\cite{yin2017knowledge}:
(i) Uniform Experience Replay (UER): The default baseline sampling transitions uniformly;
(ii) Prioritized Experience Replay (PER)~\cite{2016Prioritized}: A strategy that reweights samples based on temporal difference (TD) errors;
(iii) Combined Experience Replay (CER)~\cite{2017A}: A method that balances uniform and prioritized sampling to mitigate bias; 
and (iv) N-Step Replay~\cite{zhang2024efficient}: A technique incorporating multi-step returns to bridge TD updates with Monte Carlo estimation.

\begin{figure*}[!t]
\centering

\begin{subfigure}[b]{.24\textwidth}
  \centering
  \includegraphics[width=\linewidth]{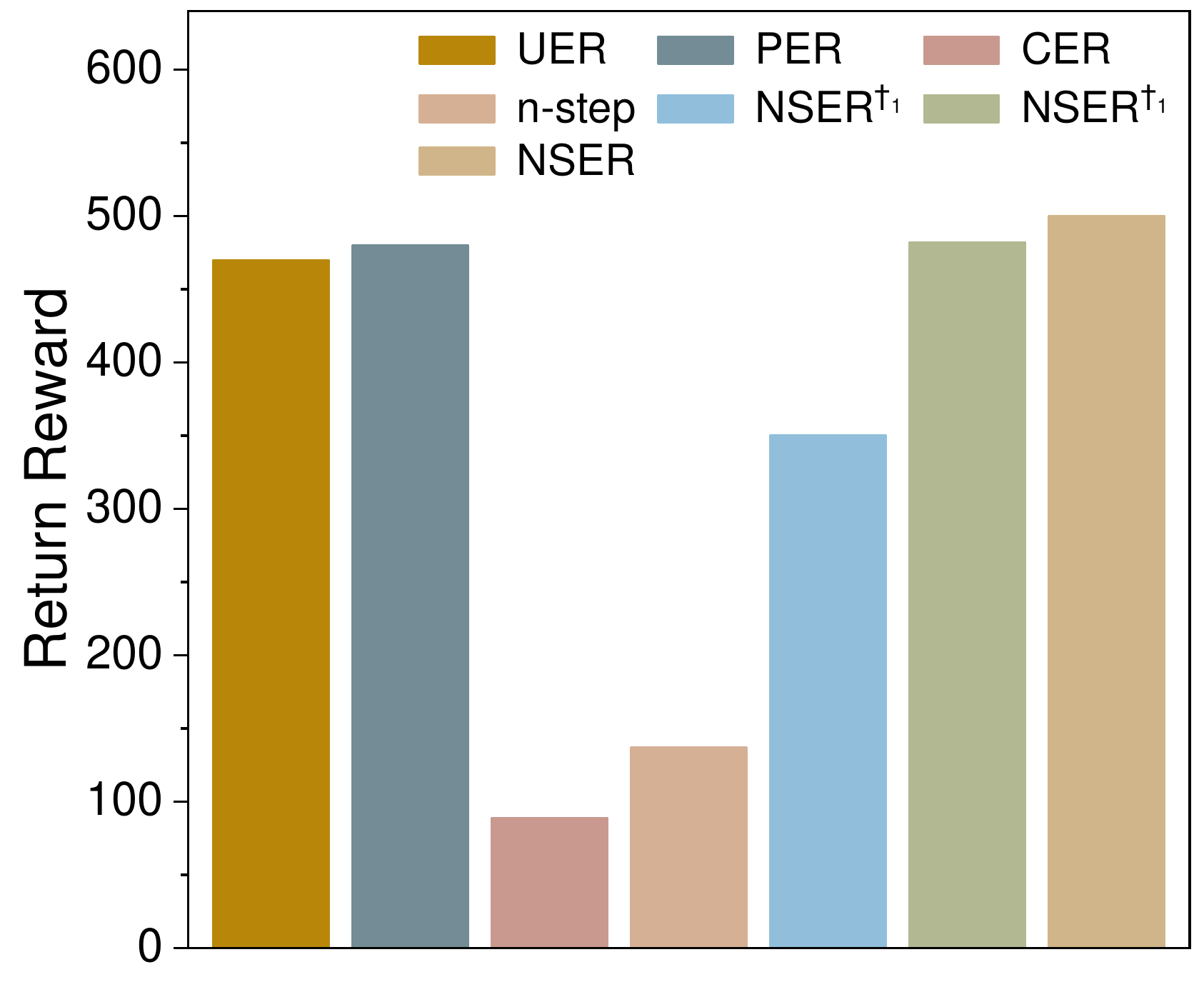}
  \subcaption*{\scriptsize (a) Prompt design: return reward}
\end{subfigure}
\hfill
\begin{subfigure}[b]{.24\textwidth}
  \centering
  \includegraphics[width=\linewidth]{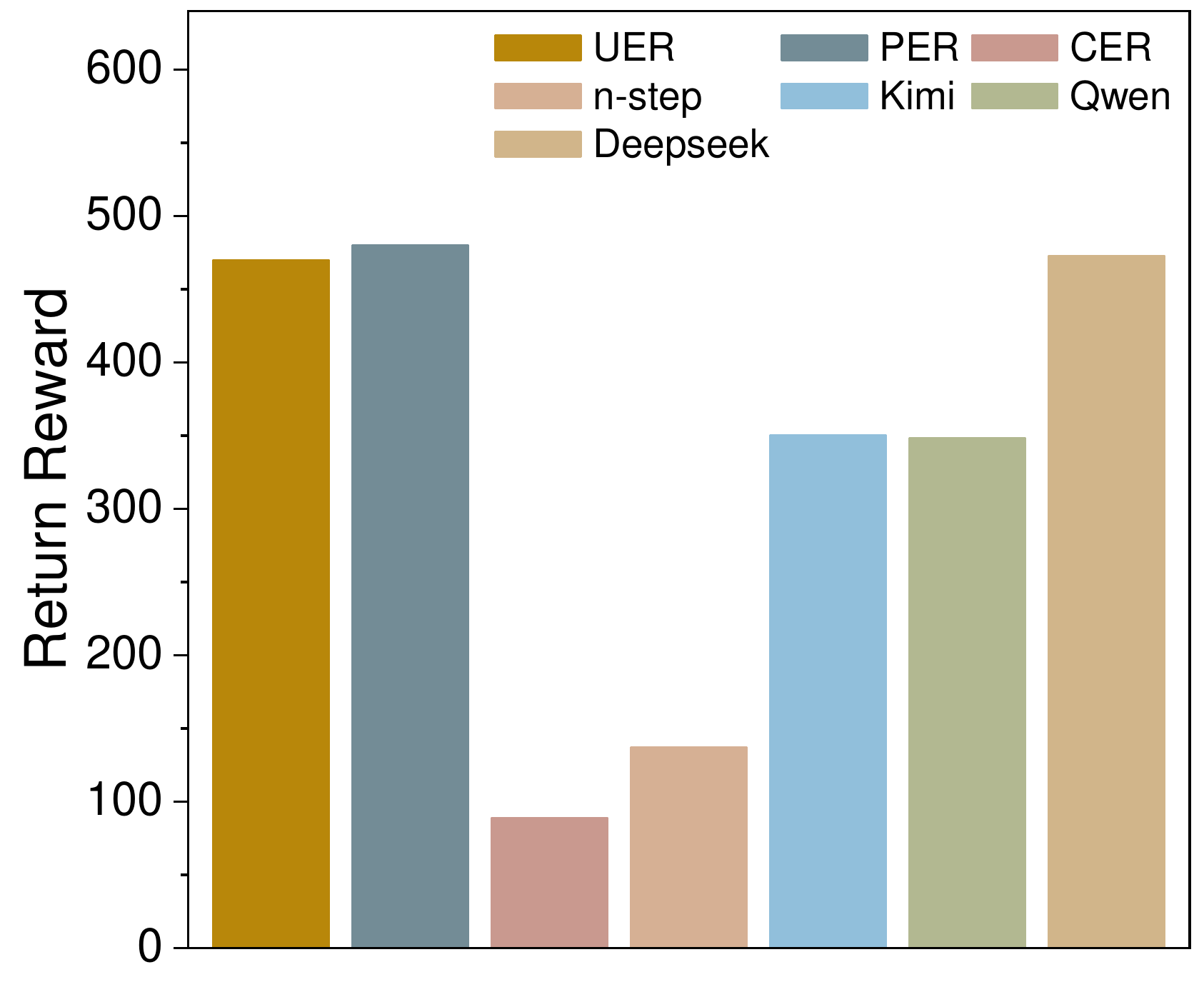}
  \subcaption*{\scriptsize (b) LLM backbone: return reward}
\end{subfigure}
\hfill
\begin{subfigure}[b]{.24\textwidth}
  \centering
  \includegraphics[width=\linewidth]{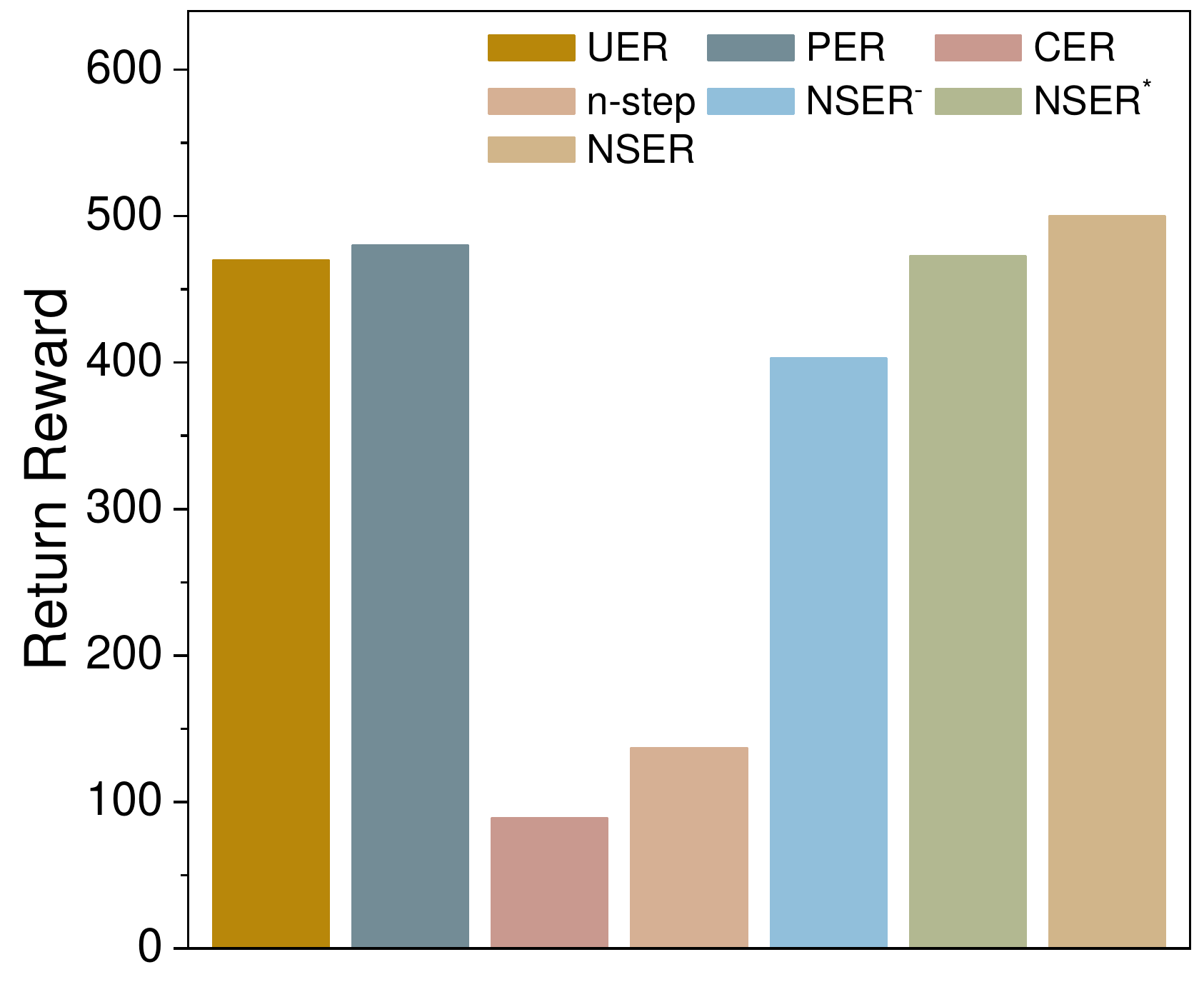}
  \subcaption*{\scriptsize (c) symbolic grounding: return reward}
\end{subfigure}
\begin{subfigure}[b]{.24\textwidth}
  \centering
  \includegraphics[width=\linewidth]{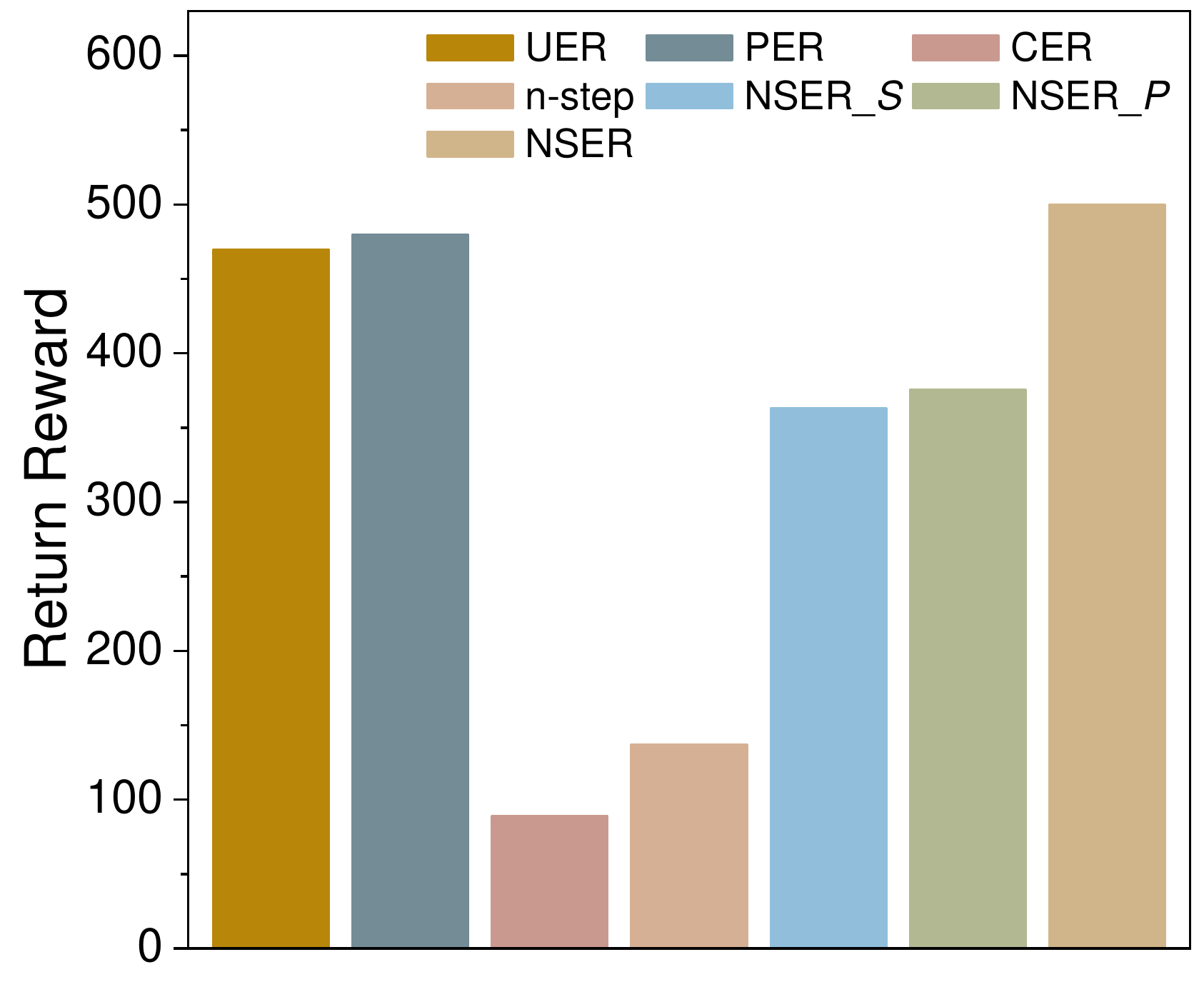}
  \subcaption*{\scriptsize (d) behavior sampling: return reward}
\end{subfigure}
% \\
% \begin{subfigure}[b]{.24\textwidth}
%   \centering
%   \includegraphics[width=\linewidth]{Figures/figa11.pdf}
%   \subcaption*{\scriptsize (e) Prompt design: training curves}
% \end{subfigure}
% \hfill
% \begin{subfigure}[b]{.24\textwidth}
%   \centering
%   \includegraphics[width=\linewidth]{Figures/figa12.pdf}
%   \subcaption*{\scriptsize (f) LLM backbone: training curves}
% \end{subfigure}
% \hfill
% \begin{subfigure}[b]{.24\textwidth}
%   \centering
%   \includegraphics[width=\linewidth]{Figures/figa13.pdf}
%   \subcaption*{\scriptsize (g) symbolic grounding: training curves}
% \end{subfigure}
% \begin{subfigure}[b]{.24\textwidth}
%   \centering
%   \includegraphics[width=\linewidth]{Figures/figa14.pdf}
%   \subcaption*{\scriptsize (h) behavior sampling: training curves}
% \end{subfigure}

\caption{
Ablation studies of NSER across various design configurations. 
Results report the final episodic returns, demonstrating that simplifying or removing individual components consistently degrades performance. 
These findings highlight the critical contributions of language-based rule induction, neuro-symbolic grounding, and behavior-guided sampling to the overall framework efficacy.
}
\label{fig:ablation}
\end{figure*}

\subsection{Hyperparameter Setting}
All methods follow standard hyperparameter settings consistent with prior work. Unless otherwise specified, all reinforcement learning components remain identical across compared methods within each environment. For NSER, we adopt the DeepSeek-v3.2 LLM~\cite{deng2025exploring} as the semantic reasoning backbone, which is used exclusively for zero-shot behavior induction and remains frozen throughout training. Symbolic relations are updated periodically from replay trajectories, while the policy learning objective remains unchanged. Detailed configuration parameters and schedules are provided in Appendix~\ref{app:b}.

\subsection{Evaluation Metrics}
We evaluate all methods from three complementary perspectives. Task performance is measured by the expected episodic return and the area under the learning curve (AUC), reflecting cumulative training performance. 
Learning efficiency is assessed by the environment interaction steps required to reach a predefined performance threshold (Steps-to-$\tau$) and the convergence step $N_{\mathrm{conv}}$. Computational overhead is evaluated using wall-clock training time, reported as the relative speedup ratio $\eta = T_{\mathrm{UER}} / T_{\mathrm{NSER}}$ over the baseline. Detailed metrics are provided in Appendix~\ref{app:c_res}.

\subsection{Algorithm and Environment Generality (Q1)}
Table~\ref{tab:1} summarizes the performance comparison across multiple off-policy RL algorithms and environment categories. 
Specifically, NSER consistently improves both episodic return and AUC across DQN, C51, and QR-DQN backbones, suggesting that the proposed knowledge-guided mechanism generalizes effectively across different value representation schemes. Significant performance gains are observed in reactive, rule-based, and procedurally generated tasks, especially those involving sparse or delayed rewards. Overall, these results underscore the robust performance and broad applicability of NSER across a diverse spectrum of RL algorithms and task domains.

\subsection{Comparison with Replay Strategies (Q2)}
We evaluate NSER against representative experience replay strategies within the DQN framework, with results summarized in Table~\ref{tab:2}. Unlike transition-level methods that prioritize experiences via local numerical signals like TD-error, NSER operates at the trajectory level by exploiting induced behavioral structures. This semantic abstraction enables NSER to capture long-horizon dependencies more effectively, yielding superior sample efficiency and enhanced convergence stability. Notably, NSER achieves these gains while maintaining a favorable performance-efficiency trade-off, significantly reducing the total interaction steps required for convergence across all benchmarks.

\begin{figure*}[!t]
\centering
\includegraphics[width=0.9\textwidth]{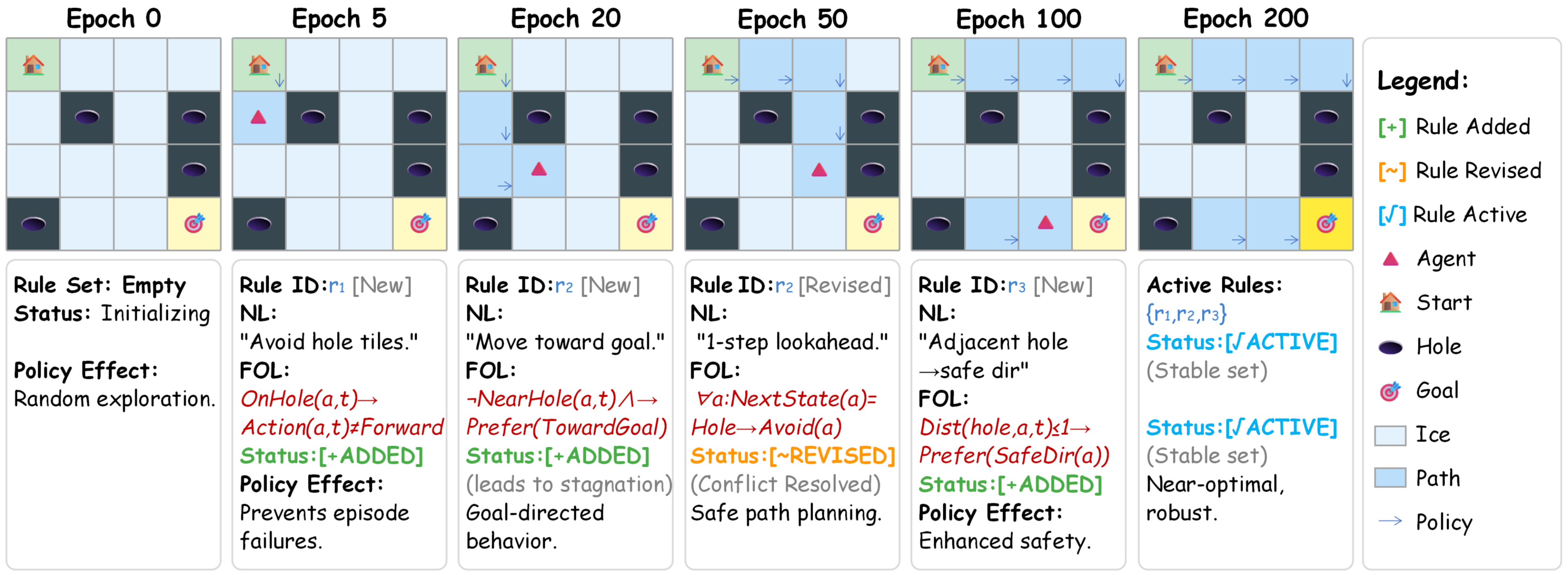}

\caption{
Temporal evolution of the induced rule set in the FrozenLake-v1 environment.
NSER initially explores without prior rules, then incrementally adds and revises behavioral rules based on accumulated experience.
Over training, the rule set converges to a stable configuration that encodes meaningful action constraints and safety preferences, resulting in consistent and robust policy behavior.
}
\label{fig:case}
\end{figure*}

\subsection{The Analysis of Active Rule Induction (Q3)}
As shown in Figure~\ref{fig:ablation}(a), we examine how prompt design affects experience understanding by comparing NSER against two variants: 
(i) NSER$^{\dagger_1}$ using a generic prompt without decision-relevant guidance; 
and (ii) NSER$^{\dagger_2}$ highlighting salient events without relating actions to outcomes. Both variants yield weakly informative descriptions, resulting in limited performance gains. 
In contrast, NSER's decision-oriented prompt emphasizes state-action-outcome transitions, enabling more interpretation and faster convergence. 
Furthermore, Figure~\ref{fig:ablation}(b) shows that while performance scales with LLM capacity, NSER consistently improves learning across different backbones, demonstrating its framework-level robustness.

\subsection{The Analysis of Neuro-Symbolic Grounding (Q4)}
As shown in Figure~\ref{fig:ablation}(c), we evaluate the necessity of differentiable grounding via two variants: (i) NSER$^{-}$ using fixed heuristic signals derived from language; and (ii) NSER$^{*}$ replacing differentiable logic with rigid \textit{if--then} rules. Both variants suffer from early performance saturation or unstable learning. In contrast, the full NSER grounds induced patterns into executable representations that undergo continuous refinement, ensuring stable convergence toward optimal behavioral policies in complex decision-making tasks.

\subsection{The Analysis of Knowledge-Guided Sampling (Q5)}
We investigate how the structural integrity of replay data supports knowledge-guided sampling by comparing different levels of interaction context. Specifically, we evaluate NSER against two variants that restrict the observational scope: (i) NSER$\_S$ utilizes isolated transitions, and (ii) NSER$\_P$ provides partial interaction trajectories. As shown in Figure~\ref{fig:ablation}(d), performance improves monotonically with increasing context completeness. This confirms that NSER's advantage stems from providing sufficient interaction context to the LLM during the replay process, rather than from simply increasing the update frequency or batch size.

\subsection{Interpretability and Evolution (Q6)}
As illustrated in Figure~\ref{fig:case}, we visualize the symbolic patterns induced during a grid-based navigation task. Rule discovery dominates early training and gradually stabilizes as the agent converges, effectively abstracting raw experience into interpretable regularities such as ``avoiding holes" and ``preferring safe directions". 
The transition from linguistic proposals to FOL ensures that this knowledge remains structured and verifiable, providing a transparent blueprint of the agent’s decision logic. 
This evolutionary process confirms that NSER progressively abstracts raw experience into robust, interpretable regularities that effectively steer the policy gradient toward long-term goals.

\section{Related Work}
In this section, we review related work on experience replay, neuro-symbolic reinforcement learning, and knowledge induction with large language models.

\subsection{Experience Replay in Reinforcement Learning}
Experience replay improves data efficiency through sample reuse, a cornerstone of off-policy reinforcement learning. Most existing strategies prioritize transitions based on local numerical signals, such as TD-errors~\cite{guan2024temporal} or short-horizon returns~\cite{gao2023stockformer}. While effective, these transition-level methods rely on short-term criteria, limiting their ability to capture long-term behavioral regularities~\cite{ghasemipour2022why}. In contrast, NSER treats trajectories as knowledge-bearing units, extracting reusable behavioral patterns from accumulated experience to guide more effective sample prioritization.

\subsection{Neuro-Symbolic Reinforcement Learning.}
Neuro-symbolic RL integrates neural networks with symbolic reasoning to enhance interpretability and generalization~\cite{Zheng2022OnlineDT}. 
Prior work typically relies on predefined symbolic spaces or task-specific rule templates, and these symbolic components are often decoupled from the experience replay process~\cite{Jin_Huang_Zhang_Hu_Nan_Du_Guo_Chen_2023}. 
Such designs restrict adaptability as new interaction data is collected~\cite{guo2023efficient}. 
In contrast, NSER leverages LLMs to distill behavioral patterns from trajectories into an evolving knowledge base. 
By grounding these patterns into a differentiable objective, NSER enables logical to guide the policy gradient directly, ensuring the symbolic layer remains adaptive and intrinsically coupled with the learning process.

\subsection{Large Language Models for Knowledge Induction}
Large language models have shown strong capabilities in abstraction and pattern
induction from data~\cite{yao2022react}.
Existing approaches often embed language models into the decision-making loop,
using them for planning or action generation, which raises stability and
efficiency challenges~\cite{li2022pre}.
Unlike prior work that embeds language models into the decision-making loop,
NSER uses language models only during replay to interpret past interactions.
The language model is never queried during environment interaction or policy execution.
This design enables knowledge induction from experience while avoiding the
stability and efficiency issues associated with online LLM-based decision
making precess.
\section{Conclusion}
This paper presents NSER, a neuro-symbolic experience replay framework that transforms passive data storage into an active training engine. By integrating large language models for active rule induction and differentiable logic for neuro-symbolic grounding, NSER enables agents to exploit structural regularities within their interaction history. Our experimental results across diverse benchmarks demonstrate that knowledge-guided sampling significantly enhances sample efficiency, training stability, and policy interpretability compared to traditional numerical replay strategies. Future work will explore the extension of NSER to multi-agent systems and increasingly complex long-horizon tasks where symbolic guidance is most critical for success.

\section*{Impact Statement}

This work investigates experience replay mechanisms in reinforcement learning
and proposes a neuro-symbolic framework to improve the reuse of past experience.
The potential impact of this study lies in advancing data-efficient learning
methods for sequential decision-making problems.
As with existing reinforcement learning techniques, inappropriate deployment
in real-world systems may lead to unintended behaviors.
Beyond the general risks commonly associated with reinforcement learning,
this work does not introduce additional direct negative societal impacts.
\bibliographystyle{abbrvnat}
\bibliography{2026ICML}
\clearpage
\newpage

\onecolumn

\section*{Appendix}

\appendix
\section{Detailed Benchmark Environments}\label{app:env}

This appendix provides detailed descriptions of all benchmark environments used in our experiments, including task objectives, state/action interfaces, reward structures, termination conditions, and the key behavioral regularities relevant to neuro-symbolic experience replay.

\subsection{Reactive Environments}\label{app:env_reactive}

\paragraph{CartPole-v1.}
CartPole is a classic control task where an agent applies left/right forces to a cart to keep a pole upright.
The observation is a low-dimensional continuous vector capturing cart position/velocity and pole angle/angular velocity, and the action space is discrete with two actions (push left/right).
A reward of $+1$ is given at each step the pole remains within the angle and position thresholds, and an episode terminates upon failure or when reaching the maximum time limit.
This environment exhibits stationary dynamics and is largely solvable with short-horizon reactive control, making it a representative reactive benchmark to evaluate learning efficiency.

\paragraph{Acrobot-v1.}
Acrobot is an underactuated two-link pendulum control task where the agent must swing the end-effector above a target height.
The observation is a continuous state vector encoding joint angles and angular velocities, and the action space is discrete (torque applied at the actuated joint).
The environment typically provides a step penalty until the goal is reached, and an episode terminates when the height threshold is achieved or when the time limit is reached.
Although success requires coordinated dynamics, the task remains stationary and feedback is immediate at the episode level, serving as a reactive benchmark for sample efficiency and stability.

\subsection{Rule-Centric Environments}\label{app:env_rulecentric}

\paragraph{FrozenLake-v1.}
FrozenLake is a gridworld navigation task where the agent starts from a fixed position and aims to reach a goal while avoiding holes.
The state is a discrete grid cell index and actions are discrete moves (up/down/left/right).
The reward is sparse (typically $1$ upon reaching the goal and $0$ otherwise), with termination upon reaching the goal or falling into a hole.
This environment contains implicit prerequisite constraints for success (e.g., avoiding unsafe states and following safe corridors), and locally appealing moves can lead to irreversible failure.
Such implicit logical structure makes FrozenLake a rule-centric benchmark for evaluating whether induced behavioral patterns improve long-horizon planning under sparse feedback.

\paragraph{Taxi-v3.}
Taxi is a symbolic gridworld task in which an agent must navigate to pick up a passenger at a designated location and drop them off at a target destination.
The state is discrete and factorized by taxi position, passenger location/status, and destination, and the action space includes navigation actions plus \texttt{pickup} and \texttt{dropoff}.
Rewards are shaped with step penalties and large positive/negative signals for successful/invalid pickup/dropoff operations, and episodes terminate after successful delivery or when the step limit is reached.
The task exhibits explicit prerequisite relations (pickup must occur before dropoff, and both require satisfying location constraints), making it strongly rule-centric and well-suited to test whether symbolic regularities extracted from replay can be reused to improve learning.

\subsection{Procedural Environments}\label{app:env_procedural}

\paragraph{Procgen-CoinRun.}
CoinRun is a procedurally generated platformer environment where the agent must traverse a level to collect a coin while avoiding hazards and obstacles.
Observations are high-dimensional (pixel-based) and actions are discrete control commands.
Rewards are sparse and primarily associated with reaching the coin (and optionally intermediate signals depending on configuration), with termination upon success, failure, or time limit.
The environment generates non-overlapping instances between training and evaluation by construction, emphasizing generalization across unseen but structurally similar levels.
This makes CoinRun a procedural benchmark to assess whether abstract behavioral knowledge can transfer beyond memorizing specific trajectories.

\paragraph{Procgen-Maze.}
Procgen-Maze is a procedurally generated navigation task where the agent must explore and reach a goal in randomly generated mazes.
Observations are typically pixel-based and actions are discrete.
Rewards are sparse and tied to reaching the goal, with termination upon success or time limit.
Because training and evaluation mazes are disjoint, performance depends on extracting reusable navigation structure (e.g., exploration and goal-directed behaviors) that generalizes to unseen layouts.
This environment serves as a procedural benchmark focusing on out-of-distribution generalization enabled by structure-aware replay.

\paragraph{Implementation Notes.}
Unless otherwise stated, we follow the default environment settings provided by the benchmark suites and use identical training budgets and random seeds across all methods to ensure fair comparison.

\begin{figure*}[!t]
    \centering
    \includegraphics[width=0.75\textwidth]{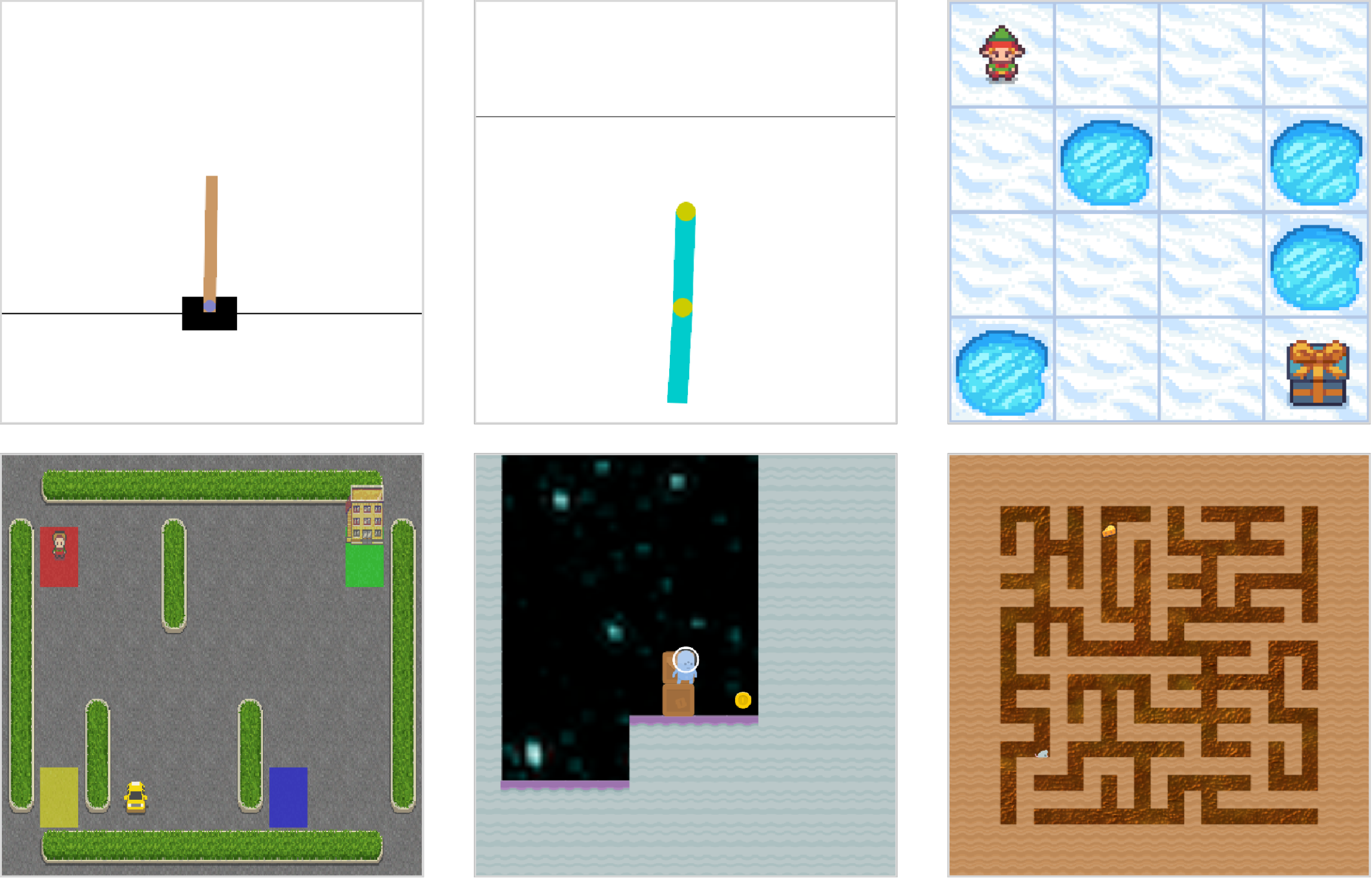}
    \caption{
    Screen shots from six benchmark environments: \textit{(from left to right, top to bottom)} CartPole-v1, Acrobot-v1, FrozenLake-v1, Taxi-v3, Procgen-CoinRun, and Procgen-Maze.
    }
    \label{fig:case_full}
\end{figure*}

\section{Implementation Details and Hyperparameters}\label{app:b}

\subsection{Reinforcement Learning Configuration}

For all environments, we use a two-layer MLP with ReLU activations as the Q-network.
The learning rate is set to $1\times10^{-3}$, the replay buffer size is $10^6$,
and the batch size is 64.
Target networks are updated every 1,000 environment steps.
An $\epsilon$-greedy exploration strategy is adopted with linear decay.

\subsection{Language Model Configuration}

The symbolic behavior induction module is instantiated using one of the following large language models: DeepSeek-V3.2, Qwen-Plus, and Kimi-Thinking-K2. These models are employed exclusively for trajectory-level behavior summarization and rule induction, and are never involved in online policy execution, action selection, or value function approximation.

All language models are used in a zero-shot setting with fixed prompts and decoding configurations. No task-specific fine-tuning or gradient-based updates are performed on the language models during reinforcement learning. The choice of language model does not affect the structure of NSER or its interaction with the underlying reinforcement learning algorithm, and the same replay prioritization mechanism is applied regardless of the specific language model used.

To limit computational overhead, the language model is invoked periodically rather than at every training step. At each induction interval, a subset of trajectories is sampled from the replay buffer and serialized into structured textual summaries that preserve the temporal order of states, actions, and rewards. The language model is queried in a zero-shot manner to generate candidate behavioral descriptions and remains frozen throughout training.

Unless otherwise stated, experiments reported in the main text use a single language model configuration for consistency, while alternative models are evaluated to verify the observed trends are not sensitive to a particular language model choice.

\subsection{Symbolic Module Configuration}

The number of relation prototypes is set to $K=16$.
The temperature parameter $\beta$ and replay prioritization coefficient $\eta$
are selected from $\{0.1, 0.5, 1.0\}$ based on validation performance and then
fixed for all experiments.

Symbolic rules induced by the language model are updated asynchronously with respect to policy learning. After each induction phase, newly generated or refined rules are integrated into the neuro-symbolic module, where their parameters are updated using accumulated replay data. Between two consecutive updates, the symbolic module is kept fixed to ensure stable replay prioritization during minibatch optimization.

\subsection{Evaluation Metrics}

We evaluate all methods from three aspects: task performance,
learning efficiency, and computational overhead.

\paragraph{Task Performance.}
Task performance is measured by the expected episodic return
\begin{equation}
\bar{R}
=
\mathbb{E}_{\pi}\!\left[ \sum_{t=0}^{T} r_t \right],
\end{equation}
estimated over multiple evaluation episodes.
We additionally report the area under the learning curve (AUC),
which reflects cumulative performance throughout training.

\paragraph{Learning Efficiency.}
Learning efficiency is quantified by the number of environment interaction
steps required to reach a predefined performance threshold $\tau$:
\begin{equation}
\text{Steps-to-}\tau
=
\min \left\{ N \mid \bar{R}_N \ge \tau \right\},
\end{equation}
where this metric directly captures sample efficiency and convergence speed.

\paragraph{Computational Overhead.}
To evaluate computational efficiency, we measure wall-clock training time and
report the relative speedup:
\begin{equation}
\eta
=
\frac{T_{\text{Baseline}}}{T_{\text{Method}}},
\end{equation}
where $\eta>1$ indicates lower computational cost compared to the baseline.

\subsection{Experimental Protocol and Parameter Consistency}

All methods are trained under equivalent training budgets. Specifically, each method is allocated the same total amount of environment interaction, corresponding to a fixed number of training episodes or an equivalent number of environment steps, depending on the characteristics of the environment. This ensures that performance comparisons are not influenced by differences in episode length or termination dynamics across tasks.

Identical network architectures, optimizer configurations, replay buffer capacities, and random seeds are used across all compared methods within each environment. No environment-specific hyperparameter tuning is performed for individual replay strategies, and all methods follow the same training and evaluation schedules.

Reported training time and efficiency metrics reflect the full end-to-end computational cost of learning, including both standard reinforcement learning updates and the additional overhead introduced by language-based trajectory analysis and symbolic rule induction. All experiments are conducted under the same hardware configuration and software environment to ensure that observed performance differences arise from the replay strategy itself rather than discrepancies in experimental setup or resource allocation.

\begin{table*}[!t]
\centering
\small
\setlength{\tabcolsep}{6pt}
\renewcommand{\arraystretch}{1.10}
\caption{
Performance of different experience replay strategies across various game environments.
All speedup values ($\eta$) are reported as training-time ratios relative to uniform experience replay (UER).
\textbf{Bold} indicates the best result in each column per environment.
}
\begin{tabular}{c c c c c c}
\toprule
\multirow{2}{*}{Game/Env} &
\multicolumn{2}{c}{Performance} &
\multicolumn{2}{c}{Efficiency} &
Overhead \\
\cmidrule(lr){2-3}\cmidrule(lr){4-5}\cmidrule(lr){6-6}
& Return $\uparrow$ & AUC $\uparrow$ &
Steps-to-$\tau$ $\downarrow$ & $N_{\mathrm{conv}}$ $\downarrow$ &
$\eta = T_{\mathrm{UER}}/T \uparrow$ \\
\midrule
Pong &
\makecell[l]{UER: $-19.6\pm1.5$\\PER: $-20.4\pm1.0$\\CER: $-12.8\pm9.7$\\NSER: $\mathbf{-7.9\pm9.6}$} &
\makecell[l]{UER: $-9.77\times10^{3}$\\PER: $-2.03\times10^{4}$\\CER: $-1.28\times10^{4}$\\NSER: $\mathbf{-1.56\times10^{5}}$} &
\makecell[l]{UER: $\mathbf{1.35\times10^{5}}$\\PER: $1.93\times10^{6}$\\CER: $1.55\times10^{6}$\\NSER: $1.63\times10^{7}$} &
\makecell[l]{UER: $\mathbf{5.34\times10^{5}}$\\PER: $9.15\times10^{5}$\\CER: $1.62\times10^{6}$\\NSER: $1.67\times10^{7}$} &
\makecell[l]{UER: 1.00\\PER: 0.57\\CER: 0.32\\NSER: 0.30} \\
\midrule
Breakout &
\makecell[l]{UER: $3.8\pm2.7$\\PER: $1.0\pm1.2$\\CER: $4.0\pm3.0$\\NSER: $\mathbf{9.8\pm8.7}$} &
\makecell[l]{UER: $1.91\times10^{3}$\\PER: $9.84\times10^{2}$\\CER: $3.99\times10^{3}$\\NSER: $\mathbf{1.03\times10^{5}}$} &
\makecell[l]{UER: $7.49\times10^{7}$\\PER: $8.71\times10^{7}$\\CER: $1.03\times10^{8}$\\NSER: $\mathbf{1.05\times10^{6}}$} &
\makecell[l]{UER: $\mathbf{1.39\times10^{5}}$\\PER: $1.65\times10^{5}$\\CER: $2.78\times10^{5}$\\NSER: $4.64\times10^{6}$} &
\makecell[l]{UER: 1.00\\PER: 0.80\\CER: 0.45\\NSER: 0.20} \\
\midrule
SpaceInvaders &
\makecell[l]{UER: $166\pm128$\\PER: $165\pm121$\\CER: $176\pm119$\\NSER: $\mathbf{188\pm123}$} &
\makecell[l]{UER: $8.27\times10^{4}$\\PER: $1.65\times10^{5}$\\CER: $1.76\times10^{5}$\\NSER: $\mathbf{5.62\times10^{6}}$} &
\makecell[l]{UER: $1.40\times10^{3}$\\PER: $3.21\times10^{3}$\\CER: $5.10\times10^{3}$\\NSER: $\mathbf{9.25\times10^{2}}$} &
\makecell[l]{UER: $2.69\times10^{5}$\\PER: $5.46\times10^{5}$\\CER: $5.69\times10^{5}$\\NSER: $\mathbf{1.78\times10^{5}}$} &
\makecell[l]{UER: 1.00\\PER: 0.44\\CER: 0.44\\NSER: \textbf{12.18}} \\
\midrule
Seaquest &
\makecell[l]{UER: $230\pm129$\\PER: $173\pm102$\\CER: $166\pm95$\\NSER: $\mathbf{264\pm146}$} &
\makecell[l]{UER: $1.15\times10^{5}$\\PER: $1.73\times10^{5}$\\CER: $1.65\times10^{5}$\\NSER: $\mathbf{1.85\times10^{6}}$} &
\makecell[l]{UER: $2.50\times10^{5}$\\PER: $3.72\times10^{5}$\\CER: $4.62\times10^{5}$\\NSER: $\mathbf{1.83\times10^{5}}$} &
\makecell[l]{UER: $4.21\times10^{5}$\\PER: $7.31\times10^{5}$\\CER: $7.29\times10^{5}$\\NSER: $\mathbf{1.11\times10^{5}}$} &
\makecell[l]{UER: 1.00\\PER: 0.54\\CER: 0.56\\NSER: 0.33} \\
\midrule
MsPacman &
\makecell[l]{UER: $479\pm334$\\PER: $428\pm402$\\CER: $473\pm350$\\NSER: $\mathbf{508\pm337}$} &
\makecell[l]{UER: $2.39\times10^{5}$\\PER: $5.27\times10^{5}$\\CER: $4.72\times10^{5}$\\NSER: $\mathbf{5.33\times10^{6}}$} &
\makecell[l]{UER: $\mathbf{4.90\times10^{5}}$\\PER: $5.08\times10^{5}$\\CER: $9.27\times10^{5}$\\NSER: $5.12\times10^{6}$} &
\makecell[l]{UER: $\mathbf{2.78\times10^{5}}$\\PER: $5.82\times10^{5}$\\CER: $5.52\times10^{5}$\\NSER: $6.14\times10^{6}$} &
\makecell[l]{UER: 1.00\\PER: 0.90\\CER: 0.95\\NSER: 0.37} \\
\bottomrule
\end{tabular}
\label{tab:merged}
\end{table*}

\section{Additional Experimental Results}\label{app:c_res}

\subsection{Implementation Details of NSER}

We provide further clarification on the implementation of the proposed NSER framework.

\paragraph{Trajectory construction and storage.}
Although policy optimization is performed at the transition level, NSER operates on complete interaction trajectories during replay analysis. A trajectory corresponds to a full episode rollout terminated by either task completion or environment-defined termination conditions. Each trajectory is stored in the replay buffer together with episode-level metadata, including cumulative return, trajectory length, and terminal outcome. When the buffer reaches its maximum capacity, newly collected trajectories replace the oldest ones following a sliding-window strategy.

\paragraph{Replay prioritization.}
Trajectory-level scores produced by the neuro-symbolic module define a behavior-guided replay distribution over stored experiences.
Sampling is performed at the trajectory level to capture long-horizon behavioral structure, after which sampled trajectories are decomposed into individual transitions for standard minibatch updates.Importantly, trajectory-level prioritization affects only the data selection process, while parameter updates remain transition-based and follow the original reinforcement learning procedure.
This design enables NSER to incorporate long-horizon behavioral information without modifying the underlying reinforcement learning algorithm, loss function, or optimization objective.

\subsection{Extended Quantitative Results}

In addition to the summarized results reported in Tables~\ref{tab:1} and~\ref{tab:2}, we provide exhaustive quantitative results to present a complete view of model performance. These include full performance tables covering all evaluated environments and replay strategies, reported as mean and standard deviation over multiple random seeds.

Table~\ref{tab:merged} presents extended comparisons across a broader set of environments and experience replay methods under identical training budgets. The overall performance trends observed in Table~\ref{tab:merged} are consistent with those reported in the main text: NSER generally achieves competitive or improved task performance and learning efficiency relative to existing replay strategies. Variations in the relative speedup metric $\eta$ reflect the trade-off between the additional overhead introduced by periodic language-based analysis and the gains in sample efficiency obtained from behavior-guided replay. Importantly, across all evaluated domains, NSER does not introduce instability in convergence behavior and remains comparable to or better than baseline replay methods.

\subsection{Additional Cases on NSER }
To provide concrete illustrations of NSER's behavioral induction mechanism, we present another two representative case studies that visualize the progressive discovery of symbolic rules and behavioral patterns across training epochs.

\begin{figure*}[h]
        \centering
        \includegraphics[width=0.9\textwidth]{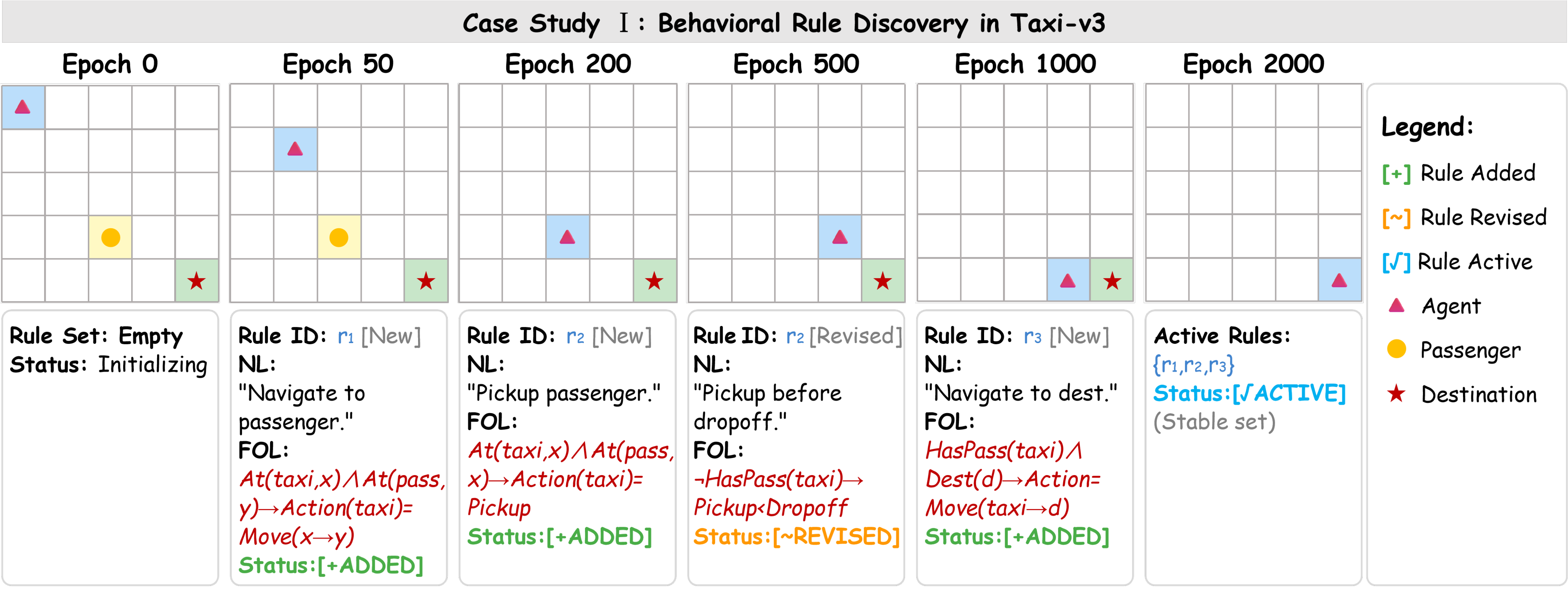}
        \caption{\textbf{Progressive rule discovery in Taxi-v3.} Training snapshots at epochs 0, 50, 200, 500, 1000, and 2000 showing (top) environment states with agent (triangle), passenger (circle), and destination (star), and (bottom) discovered rules with natural language and FOL representations. }
        \label{fig:case_study_taxi}
\end{figure*}
\begin{figure*}[t]
    \centering
    \includegraphics[width=0.9\textwidth]{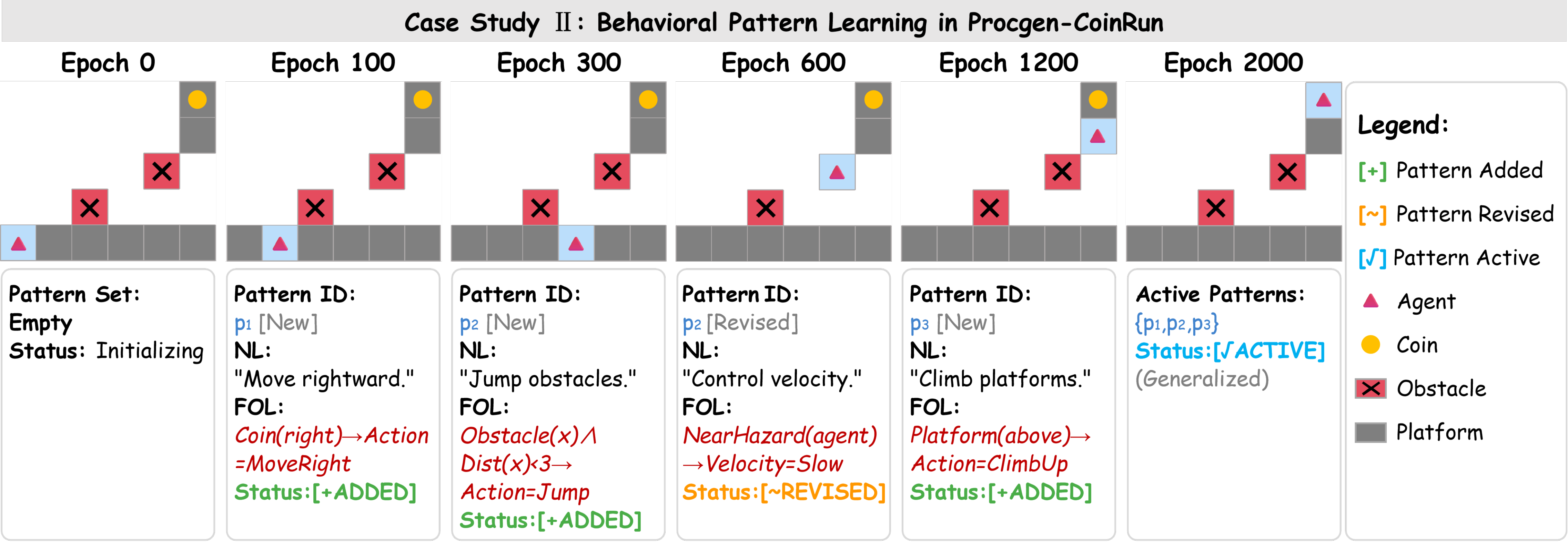}
    \caption{\textbf{Pattern learning in Procgen-CoinRun.} Training progression at epochs 0, 100, 300, 600, 1200, and 2000 showing (top) level states with agent (triangle), coin (circle), obstacles (\ding{53}), and platforms (gray), and (bottom) behavioral patterns with NL and FOL.}
    \label{fig:case_study_coinrun}
\end{figure*}

\paragraph{Case Study I: Rule Discovery in Taxi-v3.}
Figure~\ref{fig:case_study_taxi} illustrates the evolution of rule-based knowledge in the Taxi-v3 environment, a classic discrete control task requiring sequential decision-making with explicit preconditions. Each training snapshot captures both the environment state (showing agent, passenger, and destination positions) and the corresponding induced rule set with natural language and first-order logic representations. Starting from an empty rule set at initialization (Epoch 0), the language model progressively identifies navigation toward the passenger ($r_1$, Epoch 50), pickup preconditions based on location matching ($r_2$, Epoch 200), temporal ordering constraints enforcing pickup-before-dropoff dependencies (revised $r_2$, Epoch 500), and goal-directed navigation after successful pickup ($r_3$, Epoch 1000). By Epoch 2000, the rule set stabilizes into a coherent policy that robustly solves the task. The color-coded status indicators track rule evolution: green (ADDED) denotes newly discovered rules, orange (~REVISED) indicates refinement of existing rules, and blue (ACTIVE) marks convergence to a stable rule set. This case study demonstrates NSER's ability to extract interpretable, compositional knowledge structures from raw interaction experience in rule-centric domains.

\paragraph{Case Study II: Pattern Learning in Procgen-CoinRun.}
Figure~\ref{fig:case_study_coinrun} demonstrates behavioral pattern induction in Procgen-CoinRun, a procedurally generated platformer environment where generalization to unseen level layouts is critical. Unlike the discrete symbolic rules in Taxi-v3, CoinRun requires continuous control strategies that adapt to varied visual configurations. The training progression shows how the agent learns directional movement biases toward the coin ($p_1$, Epoch 100), preemptive obstacle avoidance through jumping ($p_2$, Epoch 300), refined velocity modulation near hazards (revised $p_2$, Epoch 600), and vertical platform navigation ($p_3$, Epoch 1200). By Epoch 2000, these patterns achieve level-invariant generalization, enabling robust performance on unseen procedurally generated instances. The natural language descriptions and FOL formalizations capture high-level control policies that abstract away level-specific details while preserving task-relevant behavioral structure. This case study highlights NSER's capacity to discover reusable behavioral abstractions from high-dimensional pixel observations in procedurally generated domains, bridging the gap between low-level perceptual features and high-level strategic reasoning.

% \subsection{Training Curves on Different NSER Varieties}

\section{Algorithm Pseudocode}\label{app:D_alg_pse}

This appendix provides detailed pseudocode for the proposed Neuro-Symbolic Experience Replay (NSER) framework. The algorithm consists of three main components: (1) language-based rule induction from trajectories, (2) neuro-symbolic grounding of induced rules, and (3) behavior-guided experience replay. We present the complete training procedure followed by detailed descriptions of each submodule.

\subsection{Main Training Loop}

Algorithm~\ref{alg:main} presents the overall NSER training procedure, which alternates between trajectory collection, rule induction, symbolic grounding, and policy optimization.

\begin{algorithm}[tb]
\caption{NSER Training Procedure}
\label{alg:main}
\begin{algorithmic}
\STATE {\bfseries Input:} Environment $\mathcal{E}$, initial policy $\pi_0$, replay buffer $\mathcal{D}$, LLM $p_{\text{LM}}$
\STATE {\bfseries Output:} Learned policy $\pi^*$
\STATE {\bfseries Initialize:} Relation prototypes $\{c_k\}_{k=1}^K$, symbolic predicates $\{E_P\}$
\FOR{episode $= 1$ {\bfseries to} $N$}
    \STATE \texttt{// Trajectory Collection}
    \STATE $\tau \gets \textsc{Rollout}(\mathcal{E}, \pi)$
    \STATE $\mathcal{D} \gets \mathcal{D} \cup \{\tau\}$
    \IF{episode $\bmod T_{\text{induction}} = 0$}
        \STATE \texttt{// Language Rule Induction (Section 3)}
        \STATE $\mathcal{Z} \gets \textsc{LanguageRuleInduction}(\mathcal{D}, p_{\text{LM}})$
        \STATE \texttt{// Neuro-Symbolic Grounding (Section 4)}
        \STATE $\{E_P\} \gets \textsc{SymbolicGrounding}(\mathcal{Z}, \mathcal{D})$
    \ENDIF
    \STATE \texttt{// Behavior-Guided Replay (Section 5)}
    \FOR{update\_step $= 1$ {\bfseries to} $U$}
        \STATE $\tau_{\text{batch}} \gets \textsc{SampleTrajectories}(\mathcal{D}, \{E_P\})$
        \STATE $\mathcal{B} \gets \{(s, a, r, s') \mid (s, a, r, s') \in \tau, \tau \in \tau_{\text{batch}}\}$
        \STATE $\theta \gets \theta - \alpha \nabla_\theta \mathcal{L}^{\text{RL}}(\mathcal{B}; \theta)$
    \ENDFOR
\ENDFOR
\STATE {\bfseries return} $\pi^*$
\end{algorithmic}
\end{algorithm}

\subsection{Language Rule Induction}

The language rule induction process consists of three sequential stages that progressively transform replay trajectories into structured latent behavioral relations:

\textbf{Stage 1: Trajectory Serialization.} Each trajectory $\tau = (s_0, a_0, r_0, \ldots, s_T)$ is serialized into a textual representation:
\begin{equation}
x(\tau) = [\psi_s(s_0), \psi_a(a_0), \psi_r(r_0), \ldots, \psi_s(s_T)]
\end{equation}
where $\psi_s$, $\psi_a$, $\psi_r$ are serialization functions that convert states, actions, and rewards into textual tokens.

\textbf{Stage 2: Language-Based Pattern Proposal.} For each serialized trajectory, the LLM generates $M$ candidate behavioral patterns in zero-shot mode:
\begin{equation}
u^{(m)} \sim p_{\text{LM}}(\cdot \mid x(\tau)), \quad m = 1, \ldots, M
\end{equation}
These patterns $\mathcal{U}(\tau) = \{u^{(m)}\}_{m=1}^M$ describe behavioral regularities in natural language.

\textbf{Stage 3: Relation Assignment.} Textual patterns are embedded using text encoder $\phi(\cdot)$ and aligned with learnable relation prototypes $\{c_k\}_{k=1}^K$ via similarity:
\begin{equation}
q(z_k \mid \tau) = \frac{\exp\left(\beta \max_{u \in \mathcal{U}(\tau)} \langle \phi(u), c_k \rangle\right)}{\sum_j \exp\left(\beta \max_{u \in \mathcal{U}(\tau)} \langle \phi(u), c_j \rangle\right)}
\end{equation}
The trajectory is assigned to the most likely relation: $\Phi(\tau) = \arg\max_k q(z_k \mid \tau)$. Prototypes are updated by maximizing the marginal likelihood in Equation (11).

\subsection{Neuro-Symbolic Grounding}

Algorithm~\ref{alg:grounding} describes how induced textual behavioral rules are converted into executable FOL representations with differentiable predicates. This grounding enables trajectory-level scoring based on behavioral structure.

\begin{algorithm}[tb]
  \caption{Neuro-Symbolic Grounding}
  \label{alg:grounding}
  \begin{algorithmic}
    \STATE {\bfseries Input:} Behavioral relations $\mathcal{Z}$, replay buffer $\mathcal{D}$, max iterations $N_{\text{symbolic}}$
    \STATE {\bfseries Output:} Grounded predicates $\{E_P\}$
    \STATE Initialize $\{E_P\} \gets \emptyset$
    \STATE \textit{// Symbolic Rule Construction}
    \FOR{each $z_k$ {\bfseries in} $\mathcal{Z}$}
      \STATE Parse relation: $z_k: P_1^{(k)} \land P_2^{(k)} \land \cdots \land P_{L_k}^{(k)} \Rightarrow Q_k$
      \FOR{$\ell = 1$ {\bfseries to} $L_k$}
        \IF{$E_{P_\ell^{(k)}} \notin \{E_P\}$}
          \STATE Initialize $E_{P_\ell^{(k)}}(\cdot)$ as learnable neural network
          \STATE $\{E_P\} \gets \{E_P\} \cup \{E_{P_\ell^{(k)}}\}$
        \ENDIF
      \ENDFOR
    \ENDFOR
    \STATE \textit{// Learn Predicate Parameters}
    \FOR{iteration $= 1$ {\bfseries to} $N_{\text{symbolic}}$}
      \FOR{each trajectory $\tau$ {\bfseries in} $\mathcal{D}$}
        \FOR{each predicate $P$ {\bfseries in} $\{P_\ell^{(k)}\}$}
          \STATE Compute $\mu_P(\tau) \gets \sigma(E_P(\tau))$
        \ENDFOR
        \FOR{each relation $z_k$ {\bfseries in} $\mathcal{Z}$}
          \STATE Compute $\mu_{z_k}(\tau) \gets \prod_{\ell=1}^{L_k} \mu_{P_\ell^{(k)}}(\tau)$
        \ENDFOR
      \ENDFOR
      \STATE Update $\{E_P\}$ using gradient descent on symbolic objective
    \ENDFOR
  \end{algorithmic}
\end{algorithm}

\subsection{Behavior-Guided Sampling}

Algorithm~\ref{alg:sampling} presents the trajectory sampling procedure that prioritizes replay based on behavioral structure scores. Trajectories exhibiting clearer behavioral patterns are sampled more frequently, thereby influencing policy learning through the experience distribution.

\begin{algorithm}[tb]
\caption{Behavior-Guided Trajectory Sampling}
\label{alg:sampling}
\begin{algorithmic}
\STATE {\bfseries Input:} Replay buffer $\mathcal{D}$, grounded predicates $\{E_P\}$, relations $\mathcal{Z}$
\STATE {\bfseries Output:} Sampled trajectory batch $\tau_{\text{batch}}$
\STATE Initialize $W \gets \emptyset$
\FOR{each $\tau$ {\bfseries in} $\mathcal{D}$}
    \FOR{each $z_k$ {\bfseries in} $\mathcal{Z}$}
        \STATE Compute $\mu_{z_k}(\tau) \gets \prod_{\ell=1}^{L_k} \sigma(E_{P_\ell^{(k)}}(\tau))$
    \ENDFOR
    \STATE Compute $w(\tau) \gets \sum_{k=1}^K \mu_{z_k}(\tau)^p$
    \STATE Store $W[\tau] \gets w(\tau)$
\ENDFOR
\FOR{each $\tau$ {\bfseries in} $\mathcal{D}$}
    \STATE Compute $p_{\text{replay}}(\tau) \gets \frac{\exp(\eta \cdot w(\tau))}{\sum_{\tau' \in \mathcal{D}} \exp(\eta \cdot w(\tau'))}$
\ENDFOR
\STATE Sample $\tau_{\text{batch}}$ as $B$ trajectories from $\mathcal{D}$ according to $p_{\text{replay}}$
\STATE {\bfseries return} $\tau_{\text{batch}}$
\end{algorithmic}
\end{algorithm}

\subsection{Implementation Notes}

\textbf{Trajectory Serialization ($\psi$):} The serialization functions $\psi_s$, $\psi_a$, $\psi_r$ convert states, actions, and rewards into textual tokens that preserve semantic meaning. For discrete environments, we use symbolic names (e.g., ``position=(2,3)'', ``action=move\_right''). For continuous state spaces, we discretize or use binned descriptions.

\textbf{LLM Prompting:} The zero-shot prompt provided to $p_{\text{LM}}$ explicitly instructs the model to identify decision-relevant patterns, emphasizing state transitions, action preconditions, and outcome dependencies. The prompt template is fixed across all environments and never involves task-specific fine-tuning.

\textbf{Relation Prototype Optimization:} The prototypes $\{c_k\}$ are updated using stochastic gradient ascent on the marginal likelihood in Equation (11). In practice, we use Adam optimizer with learning rate $10^{-3}$ and update prototypes every $T_{\text{induction}}$ episodes.

\textbf{Predicate Initialization:} Each predicate scoring function $E_P(\cdot)$ is implemented as a two-layer MLP with ReLU activation, mapping trajectory embeddings to scalar scores. Trajectory embeddings are constructed by mean-pooling state representations across time steps.

\textbf{Computational Efficiency:} Rule induction is performed asynchronously and does not block policy training. While the LLM processes trajectories, the RL agent continues learning using previously computed scores. This design ensures minimal overhead while maintaining training stability.

\subsection{Computational Complexity}

We analyze the computational complexity of NSER compared to standard experience replay. Let $N$ denote the replay buffer size, $K$ the number of relations, $M$ the number of candidate patterns per trajectory, and $L$ the average number of predicates per relation.

\begin{itemize}
    \item \textbf{Language Rule Induction:} $O(N \cdot M \cdot d)$, where $d$ is the text encoder embedding dimension. This occurs every $T_{\text{induction}}$ episodes and is performed offline.
    \item \textbf{Symbolic Grounding:} $O(N \cdot K \cdot L \cdot T)$, where $T$ is the trajectory length. Predicate evaluation is vectorized across batch dimensions for efficiency.
    \item \textbf{Behavior-Guided Sampling:} $O(N \cdot K \cdot L)$ per sampling operation, equivalent to computing forward passes through $K \cdot L$ predicate networks.
    \item \textbf{Baseline (Uniform Replay):} $O(N)$ for uniform sampling, with no additional overhead.
\end{itemize}

In practice, the dominant cost is LLM inference during rule induction. However, since this occurs infrequently (every $T_{\text{induction}} = 50$ episodes in our experiments) and asynchronously, the amortized overhead per training step remains low. The symbolic grounding and sampling operations add negligible cost compared to policy network updates.

\subsection{Efficiency Analysis}

We provide a detailed efficiency analysis comparing NSER with standard uniform experience replay (UER). 
Our analysis characterizes both the additional computational overhead introduced by neuro-symbolic processing and the resulting gains in sample efficiency due to behavior-guided replay.
Unless otherwise specified, all complexity terms are measured in wall-clock time.

\subsubsection{Wall-Clock Time Analysis}

The total training time of NSER can be decomposed as:
\begin{equation}
T_{\text{NSER}} = T_{\text{collection}} + T_{\text{induction}} + T_{\text{grounding}} + T_{\text{update}},
\end{equation}
where each term corresponds to a distinct stage of the training pipeline.

\paragraph{Trajectory Collection Time.}
\begin{equation}
T_{\text{collection}} = N_{\text{episodes}} \cdot \mathbb{E}_{\pi}[T_{\text{episode}}],
\end{equation}
where $N_{\text{episodes}}$ is the total number of training episodes and $T_{\text{episode}}$ denotes the episode length under policy $\pi$.
This term is identical across NSER and UER, as it depends only on environment dynamics and agent interaction.

\paragraph{Rule Induction Time.}
Rule induction is performed periodically every $T_{\text{induction}}$ episodes.
The total induction cost is:
\begin{equation}
T_{\text{induction}} =
\frac{N_{\text{episodes}}}{T_{\text{induction}}}
\cdot
\left(
T_{\text{LLM}} + T_{\text{alignment}}
\right),
\end{equation}
where:
\begin{align}
T_{\text{LLM}} &= N_{\text{sample}} \cdot M \cdot t_{\text{inference}}, \\
T_{\text{alignment}} &= N_{\text{sample}} \cdot K \cdot d \cdot t_{\text{embed}}.
\end{align}
Here, $N_{\text{sample}}$ denotes the number of sampled trajectories used for rule induction, $M$ is the number of candidate behavioral patterns generated per trajectory, and $t_{\text{inference}}$ is the per-pattern LLM inference time.
The alignment cost accounts for embedding-based matching between induced language rules and $K$ latent behavioral prototypes in a $d$-dimensional embedding space.

\paragraph{Symbolic Grounding Time.}
Symbolic grounding propagates induced behavioral relations to the replay buffer:
\begin{equation}
T_{\text{grounding}} =
\frac{N_{\text{episodes}}}{T_{\text{induction}}}
\cdot
N_{\text{symbolic}} \cdot N_{\text{buffer}} \cdot K \cdot L \cdot t_{\text{predicate}},
\end{equation}
where $N_{\text{symbolic}}$ is the number of grounding iterations, $N_{\text{buffer}}$ is the replay buffer size, $L$ is the number of predicates per rule, and $t_{\text{predicate}}$ denotes the forward-pass cost of a single predicate network.

\paragraph{Policy Update Time.}
\begin{equation}
T_{\text{update}} =
N_{\text{episodes}} \cdot U \cdot (t_{\text{sample}} + t_{\text{policy}}),
\end{equation}
where $U$ is the number of gradient updates per episode, $t_{\text{policy}}$ is the cost of a single policy update, and $t_{\text{sample}}$ denotes the replay sampling cost.
For NSER:
\begin{equation}
t_{\text{sample}}^{\text{NSER}} =
N_{\text{buffer}} \cdot K \cdot L \cdot t_{\text{score}} + B \cdot t_{\text{softmax}},
\end{equation}
where trajectory-level scores are computed prior to minibatch selection.
In contrast, uniform replay incurs:
\begin{equation}
t_{\text{sample}}^{\text{UER}} = B \cdot t_{\text{uniform}}.
\end{equation}

\paragraph{Speedup Ratio.}
Overall wall-clock efficiency is quantified by:
\begin{equation}
\eta =
\frac{T_{\text{UER}}}{T_{\text{NSER}}}
=
\frac{
T_{\text{collection}} + N_{\text{episodes}} \cdot U \cdot (t_{\text{uniform}} + t_{\text{policy}})
}{
T_{\text{collection}} + T_{\text{induction}} + T_{\text{grounding}} + N_{\text{episodes}} \cdot U \cdot (t_{\text{sample}}^{\text{NSER}} + t_{\text{policy}})
}.
\end{equation}

\subsubsection{Sample Efficiency Analysis}

NSER improves sample efficiency by prioritizing trajectories that are behaviorally informative.
Let $\epsilon_{\text{target}}$ denote a target performance threshold.
We define the sample complexity as:
\begin{equation}
N_{\text{steps}}(\epsilon) =
\min \left\{
n : \mathbb{E}_{\pi_n}[R] \ge \epsilon_{\text{target}}
\right\},
\end{equation}
where $\pi_n$ is the policy after $n$ environment steps.

Under uniform replay, convergence typically follows:
\begin{equation}
\mathbb{E}_{\pi_n}[R] \approx R^* - O\!\left(\frac{1}{\sqrt{n}}\right).
\end{equation}
With behavior-guided replay, NSER effectively reweights trajectories according to their long-horizon utility, yielding:
\begin{equation}
\mathbb{E}_{\pi_n}[R] \approx R^* - O\!\left(\frac{1}{\sqrt{n \cdot \kappa}}\right),
\end{equation}
where $\kappa > 1$ is an effective sample efficiency multiplier.

\paragraph{Convergence Acceleration Factor.}
\begin{equation}
\kappa =
\frac{
\text{Var}_{p_{\text{uniform}}(\tau)}[Q^\pi(\tau)]
}{
\text{Var}_{p_{\text{replay}}(\tau)}[Q^\pi(\tau)]
}
\cdot
\frac{
\mathbb{E}_{p_{\text{replay}}(\tau)}[w(\tau)]
}{
\mathbb{E}_{p_{\text{uniform}}(\tau)}[w(\tau)]
},
\end{equation}
where $Q^\pi(\tau)$ denotes the trajectory-level return and $w(\tau)$ is the induced importance weight.
This expression captures variance reduction and preferential emphasis on high-value trajectories.

\subsubsection{Amortized Cost Analysis}

The amortized per-update overhead of NSER is:
\begin{equation}
c_{\text{step}} =
c_{\text{LLM}}^{\text{step}} +
c_{\text{align}}^{\text{step}} +
c_{\text{ground}}^{\text{step}} +
c_{\text{score}}^{\text{step}},
\end{equation}
with:
\begin{itemize}
\item \textbf{LLM inference:}
\(
c_{\text{LLM}}^{\text{step}} =
\frac{
N_{\text{inductions}} \cdot N_{\text{sample}} \cdot M \cdot t_{\text{inference}}
}{
N \cdot U
}.
\)

\item \textbf{Prototype alignment:}
\(
c_{\text{align}}^{\text{step}} =
\frac{
N_{\text{inductions}} \cdot N_{\text{sample}} \cdot K \cdot d \cdot t_{\text{embed}}
}{
N \cdot U
}.
\)

\item \textbf{Symbolic grounding:}
\(
c_{\text{ground}}^{\text{step}} =
\frac{
N_{\text{inductions}} \cdot N_{\text{symbolic}} \cdot N_{\text{buffer}} \cdot K \cdot L \cdot t_{\text{predicate}}
}{
N \cdot U
}.
\)

\item \textbf{Trajectory scoring:}
\(
c_{\text{score}}^{\text{step}} =
N_{\text{buffer}} \cdot K \cdot L \cdot t_{\text{score}}.
\)
\end{itemize}

Since $T_{\text{induction}} \gg 1$ in practice (typically 50--100 episodes), the first three terms are amortized over many updates, rendering their contribution minor relative to $t_{\text{policy}}$.

\subsubsection{Theoretical Efficiency Bounds}

\paragraph{Proposition 1 (Overhead Bound).}
Assuming $T_{\text{induction}} \ge 50$ and $N_{\text{buffer}} \le 10^6$, the relative computational overhead satisfies:
\begin{equation}
\frac{T_{\text{NSER}}}{T_{\text{UER}}}
\le
1 +
\frac{
N_{\text{sample}} \cdot M \cdot t_{\text{inference}}
}{
T_{\text{induction}} \cdot U \cdot t_{\text{policy}}
}
+
O\!\left(
\frac{K \cdot L \cdot t_{\text{score}}}{t_{\text{policy}}}
\right).
\end{equation}

\paragraph{Proposition 2 (Sample Efficiency Bound).}
If induced behavioral relations partition trajectories according to return, then:
\begin{equation}
N_{\text{steps}}^{\text{NSER}}(\epsilon)
\le
\frac{N_{\text{steps}}^{\text{UER}}(\epsilon)}{\kappa}
+
O\!\left(
\frac{K \cdot L}{T_{\text{induction}}}
\right).
\end{equation}

\paragraph{Corollary.}
When $\kappa \ge \eta$, NSER achieves lower total training time:
\begin{equation}
T_{\text{total}}^{\text{NSER}}
\le
T_{\text{total}}^{\text{UER}}.
\end{equation}

\section{Illustrative Zero-Shot Prompting Example in FrozenLake}
\label{app:zeroshot_frozenlake}

To provide an intuitive illustration of the language-based rule induction process, we present a zero-shot prompting example in the \textbf{FrozenLake} environment shown in Figure~\ref{fig:case}.
This example complements the rule evolution analysis shown in Figure~4 and is intended solely for qualitative demonstration.
Importantly, the zero-shot prompting mechanism is not required for training NSER and does not introduce additional supervision.

\subsection{Environment and Trajectory Context}

FrozenLake is a grid-based navigation task in which an agent must reach a goal location while avoiding unsafe cells (holes).
Each episode generates a trajectory
$\tau = \{(s_t, a_t, r_t)\}_{t=1}^{T}$,
where states encode the agent position and local terrain type, actions correspond to directional movements, and rewards are sparse and delayed.

For this illustration, we sample a short trajectory from the replay buffer collected during early training, before stable symbolic rules have been formed.

\subsection{Zero-Shot Prompt}

Given a sampled trajectory, the language model is queried using the following zero-shot prompt:

\begin{quote}
\small
\textbf{Prompt:}  
``You are given a sequence of state--action--reward transitions from an agent interacting with a grid-based navigation environment.
Based only on the observed trajectory, identify any high-level behavioral patterns, constraints, or preferences that appear to govern successful behavior.
Describe each pattern as a general rule in natural language, without referring to specific state indices or coordinates.''
\end{quote}

No environment-specific annotations, task descriptions, or handcrafted rule templates are provided.

\subsection{Relation to Symbolic Rule Evolution}

The zero-shot rules above closely align with the symbolic patterns that emerge and stabilize during training, as visualized in Figure~4.
In NSER, such language-induced rules are subsequently embedded, aligned with latent behavioral prototypes, and grounded into differentiable symbolic predicates.
Over training, these predicates are refined, revised, or discarded based on accumulated experience, leading to the structured and interpretable rule set observed in later epochs.

This example demonstrates that meaningful behavioral abstractions can be extracted directly from raw trajectories in a zero-shot manner, providing an intuitive foundation for the behavior-guided replay mechanism employed by NSER.

\end{document}